\documentclass[10pt,twocolumn,letterpaper]{article}

\usepackage[pagenumbers]{cvpr}   % ArXiv (显示作者，有页码)

\PassOptionsToPackage{numbers,sort&compress}{natbib}

\usepackage{booktabs}
\usepackage{multirow}
\usepackage{makecell}
\usepackage{threeparttable}
\usepackage{graphicx}
\usepackage{amsmath}
\usepackage{capt-of}

\definecolor{cvprblue}{rgb}{0.21,0.49,0.74}
\usepackage[pagebackref,breaklinks,colorlinks,allcolors=cvprblue]{hyperref}

\title{Uni-Neur2Img: Unified Neural Signal-Guided Image Generation, Editing, and Stylization via Diffusion Transformers}

\author{
    \textbf{Xiyue Bai}$^{1}$ \quad \textbf{Ronghao Yu}$^{2}$ \quad \textbf{Jia Xiu}$^{1}$ \quad
    \textbf{Pengfei Zhou}$^{3}$ \quad \textbf{Jie Xia}$^{2}$ \quad \textbf{Peng Ji}$^{1}$ \\[0.5em] 
    $^{1}$Fudan University\quad
    $^{2}$Zhejiang University\quad
    $^{3}$National University of Singapore\\
    {\tt\small \url{https://github.com/BeverlyYue15/neur2img}}
}

\begin{document}

\twocolumn[{%
\renewcommand\twocolumn[1][]{#1}
\maketitle

\vspace{-0.6em}
\begin{center}
% 请确保图片路径正确
\includegraphics[width=\textwidth,trim=0mm 42mm 0mm 0mm,clip]{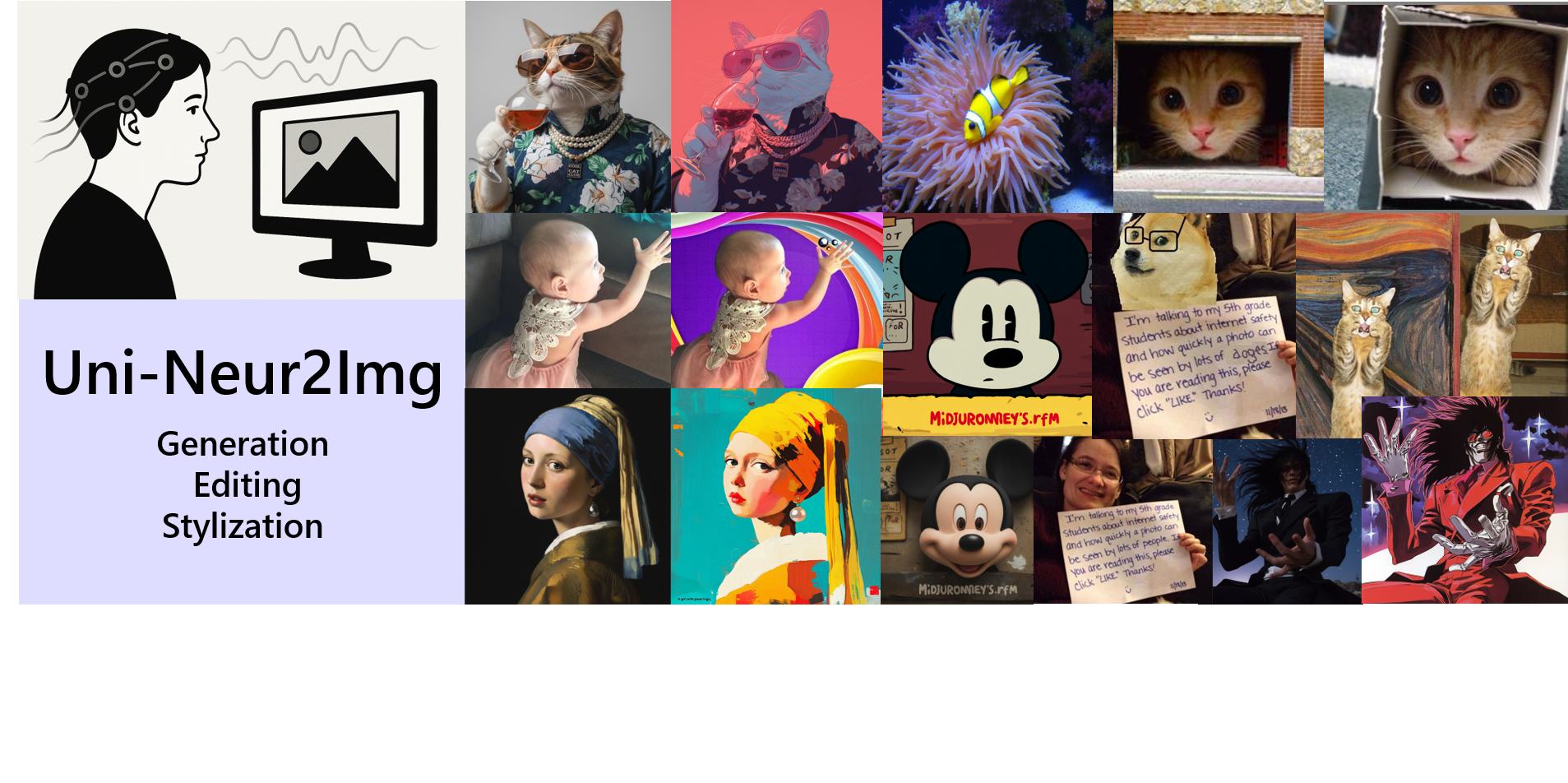}
\captionof{figure}{Illustration of Uni-Neur2Img for Image Generation, Editing, and Stylization.}
\label{fig:teaser}
\end{center}
\vspace{0.6em}
}]

\begin{abstract}
 Generating or editing images directly from Neural signals has immense potential at the intersection of neuroscience, vision, and Brain-computer interaction. In this paper, We present Uni-Neur2Img, a unified framework for neural signal-driven image generation and editing. The framework introduces a parameter-efficient LoRA-based neural signal injection module that independently processes each conditioning signal as a pluggable component, facilitating flexible multi-modal conditioning without altering base model parameters. Additionally, we employ a causal attention mechanism accommodate the long-sequence modeling demands of conditional generation tasks. Existing neural-driven generation research predominantly focuses on textual modalities as conditions or intermediate representations, resulting in limited exploration of visual modalities as direct conditioning signals. To bridge this research gap, we introduce the EEG-Style dataset. We conduct comprehensive evaluations across public benchmarks and self-collected neural signal datasets: (1) EEG-driven image generation on the public CVPR40 dataset; (2) neural signal-guided image editing on the public Loongx dataset for semantic-aware local modifications; and (3) EEG-driven style transfer on our self-collected EEG-Style dataset. Extensive experimental results demonstrate significant improvements in generation fidelity, editing consistency, and style transfer quality while maintaining low computational overhead and strong scalability to additional modalities. Thus, Uni-Neur2Img offers a unified, efficient, and extensible solution for bridging neural signals and visual content generation.

%	To bridge this research gap, we introduce EEG-Style, a new large-scale dataset specifically constructed for EEG-driven style transfer. EEG-Style contains nearly one thousand distinct visual styles and synchronized multi-subject EEG recordings acquired while participants view multiple exemplars of each style, enabling direct supervision for non-linguistic style conditioning.
	%We conduct comprehensive evaluations on three tasks: (1) EEG-driven image generation on the public CVPR40 dataset; (2) neural signal-guided image editing on the LoongX dataset; and (3) the first EEG-driven image stylization benchmark on EEG-Style, where images are stylized purely from brain activity without any textual description.
	%Surprisingly, our experiments show that EEG alone is sufficient to recover global color palettes, material appearances, and coarse stylistic biases across diverse content, revealing that low-level style statistics are explicitly reflected in non-invasive neural signals. Extensive experimental results demonstrate significant improvements in generation fidelity, editing consistency, and style transfer quality while maintaining low computational overhead and strong scalability to additional modalities. Thus, Uni-Neur2Img offers a unified, efficient, and extensible solution for bridging neural signals and visual content generation.

\end{abstract}

\section{Introduction}

The ability to generate or manipulate visual imagery directly from brain activity has long been regarded as a central ambition at the intersection of neuroscience and artificial intelligence\cite{bci1,imgrecon,avatar,bvl}. In recent years, Electroencephalography (EEG)-driven image generation and editing have emerged as a prominent research frontier across computer vision and brain--computer interface (BCI) studies\cite{eeg2visual,dreamdiffusion,eegrecons}. Envision a system capable of externalizing an artist’s imagined visual style into a concrete image, or transforming an existing photograph strictly based on mental intention. Such a capability is conceptually compelling yet technically challenging.

Early EEG-to-image translation models\cite{earlyeeg2img} were mainly based on CNN architectures such as U-Net-style autoencoders\cite{unet} and GANs\cite{gan}. However, these models typically suffer from limited consistency, weak controllability, and insufficient preservation of fine-grained visual details\cite{beatsgan}. They often produce outputs that fail to align with EEG-encoded intent, exhibit mode collapse, or lack structural fidelity. With the advent of Diffusion Transformers\cite{DiT}, generative models have acquired significantly enhanced representational capacity, more expressive conditioning mechanisms, and improved controllability\cite{controllability}, thereby opening new pathways for EEG-conditioned visual generation.

In this work, we introduce \textbf{Uni-Neur2Img} (Fig. \ref{fig:teaser}, a unified Diffusion Transformer framework that supports three EEG-conditioned tasks: (i) EEG-driven image generation, (ii) EEG-driven image editing, and (iii) EEG-driven style transfer. Methodologically, we develop a Vision--EEG encoder that compresses raw EEG time series into compact EEG tokens\cite{eegtoken}. These tokens are concatenated with image denoising tokens and jointly processed by a Diffusion Transformer equipped with cross-attention, enabling coherent and fine-grained conditional control. To further improve modularity and parameter efficiency, we propose a LoRA-based \cite{lora1} Neural Signal Injection Module that treats each conditional signal as an independent plug-and-play component. This allows multi-modal conditioning\cite{bvl} without modifying the underlying generative model. Moreover, we adopt causal attention to meet the long-sequence modeling requirements inherent in EEG-driven generation.

Among the supported tasks, \textbf{EEG-driven stylization} exhibits particularly distinctive challenges and opportunities. Existing generation and editing tasks commonly employ textual prompts as an intermediate representation for EEG signals, thereby underemphasizing the role of human visual perceptual capability\cite{insuf-l,unprompt,style1,eegtoken}. Conventional text-guided style transfer requires users to provide explicit linguistic descriptions of visual styles (e.g., ``Ghibli style'' or ``cyberpunk style''). However, most individuals are unable to precisely articulate the defining characteristics of a visual style, even though they can reliably recognize whether an image belongs to that style\cite{style2.1,style2.2}. This mismatch between perceptual style categorization and linguistic expression leads to overly abstract or imprecise prompts that limit the effectiveness of text-based style control, particularly for novel or fine-grained styles.

In contrast, the human brain acts as a powerful natural encoder and decoder of visual style\cite{brain1,brain2,brain3,brain4}. Even when individuals cannot describe a style verbally, they can readily identify it and mentally imagine how other images would appear if transformed into that style. This cognitive property motivates our exploration of non-linguistic, EEG-guided style transfer.

To enable this investigation, we construct \textbf{EEG-Style}, a new dataset comprising nearly one thousand visual styles, each represented by multiple images. We collect synchronized EEG recordings from multiple participants as they view these style exemplars, enabling precise alignment between neural responses and style categories. This dataset provides a unique foundation for studying the neural representation of visual style and, to our knowledge, EEG-Style is the first dataset that enables visual artistic stylization directly guided by EEG signals without any explicit linguistic style descriptions.

Our main contributions can be summarized as follows:
\begin{itemize}
\item  A unified Neural signal-conditioned diffusion framework that supports reconstruction, editing, and style transfer in one model.  
\item  An in-context learning based mechanism for Neural Signal-conditioned diffusion, enabling flexible adaptation to multiple tasks.  
\item  A newly collected EEG–image style-transfer dataset, along with extensive experiments proving the effectiveness and superiority of our approach.
\item Our unified framework achieves \textbf{state-of-the-art} performance across all evaluated tasks.
\end{itemize}

\section{Related works}

\subsection{Visual decoding using neural signals}

Extensive evidence shows that non-invasive neural signals—especially EEG—not only carry category-level information but also contain low-level visual attributes that can be directly decoded by machine learning methods\cite{neural1,fmri2}. In the color domain, multichannel scalp-EEG patterns can distinguish different hues and remain stable across luminance levels, suggesting that color information emerges in early temporal windows\cite{eegcolor}. In the shape–edge domain, sustained EEG potentials can decode the orientation and direction of perceived or remembered lines \cite{eegline}. In the spatial domain, alpha-band topographies can reconstruct attentional or working-memory position-channel responses through encoding models, enabling the tracking of coarse spatial locations\cite{eegapce}. These findings indicate that even with low spatial resolution, EEG can reflect key perceptual attributes such as color, orientation, and spatial position, which forms the basis for visual reconstruction.

%In contrast, fMRI provides high spatial resolution and has enabled object-category decoding of distributed voxel patterns in the visual cortex \cite{fmri1} Research has evolved from recognition to reconstruction—from binary-pattern recovery\cite{fmri2} to natural-image and movie reconstruction \cite{fmri3} and to pixel-level reconstruction using DNN features \cite{fmri5}. Recent approaches integrating latent diffusion models significantly improve reconstruction quality\cite{fmricvpr}\cite{fmricvpr2}. MEG, with millisecond temporal precision and source imaging, is often combined with fMRI for spatiotemporal fusion to map object processing dynamics\cite{fmri6}.

In recent years, several research routes have emerged for EEG-based image reconstruction\cite{eegrecons}. The first involves \text{cross-modal} alignment with diffusion priors\cite{bvl}, where EEG embeddings are aligned with CLIP or visual-semantic spaces, and two-stage or retrieval-augmented diffusion models are employed to generate images\cite{eegaaai}, achieving perceptible natural-image reconstruction under zero-shot or cross-subject settings\cite{eegnips1}. The second route focuses on end-to-end or self-supervised enhancement,
such as EEG2Video\cite{eeg2video}. Additional work proposes regional-semantic-aware EEG-LDM reconstruction \cite{eegReginalsemantic}, further improving spatial coherence and semantic alignment. Compared with fMRI\cite{fmri2} and MEG\cite{meg2image}, EEG offers millisecond temporal resolution, portability, low cost, and suitability for long-duration natural-scene recording, making it particularly advantageous for real-time or closed-loop brain–computer interfaces (BCIs)\cite{bci} and large-scale applications\cite{eegnips1}. Overall, combining EEG’s temporal dynamics and low-level perceptual cues with powerful generative priors\cite{dreamdiffusion} and alignment mechanisms represents a key direction for advancing EEG→image reconstruction toward high semantic fidelity and practical deployment.

\subsection{Image Generate model}
The mainstream path for controllable diffusion is to model conditioning signals inside the denoiser without altering the forward diffusion or the sampler. On U‑Net backbones\cite{unet}, representative approaches freeze the main model and attach lightweight external branches to ingest spatial conditions—e.g. ControlNet\cite{controlnet},  T2I‑Adapter\cite{T2i} and IP-Adapter\cite{ipadapter}, map edges/depth/segmentation/color hints to features aligned with intermediate U‑Net activations, enabling precise control over generation and editing. 

As Diffusion Transformers (DiT)\cite{peebles2023scalable} increasingly supersede U-Nets as the state-of-the-art denoiser, the conditioning framework has evolved. The DiT architecture, built upon standard Vision Transformers \cite{dosovitskiy2020image}, exposes a token-level interface that unifies text and image modalities through cross-attention\cite{chen2024pixart}. Within this framework, LoRA (Low-Rank Adaptation)\cite{incontextlora,lora1,lora2} has become a standard for activating DiT’s in-context generation capabilities, enabling rapid adaptation to downstream tasks with minimal parameter overhead\cite{huang2025photodoodle}. Building on these advancements, the FLUX\cite{flux} family integrates Rectified Flow matching \cite{liu2023flow, lipman2022flow} into the Transformer backbone, achieving superior text–image alignment and high-fidelity generation. Complementing this architecture, OminiControl\cite{ominicontrol,song2025omniconsistency} proposes a minimalist, unified conditioning pathway: rather than relying on heavy external encoders, it reuses the DiT backbone itself to encode conditional images. This approach streamlines the token-level condition injection, effectively strengthening the conditional distribution $\varepsilon_\theta(x_t,t\mid c)$ for both generation and editing tasks while maintaining compatibility with modern flow-matching denoiser.

%As Diffusion Transformers (DiT)\cite{DiT} increasingly replace U‑Nets as the denoiser, cross‑modal attention exposes a token‑level external KV interface that unifies text and image conditions and improves composability. 
%In this framework,  LoRA\cite{incontextlora,lora1,lora2} activates the DiT’s in-context generation capability with minimal additional parameters, enabling rapid adaptation to tasks such as image generation \cite{song2025omniconsistency} and image editing \cite{huang2025photodoodle}.
% Complementarily, OminiControl\cite{ominicontrol} proposes a minimal, unified image‑conditioning path: it reuses the DiT backbone to encode the conditional image 
 %it works out of the box with DiT and FLUX\cite{flux}. The FLUX family itself employs a Rectified‑Flow/flow‑matching\cite{flowmatch} Transformer denoiser with strong text–image alignment and high image fidelity, while retaining the same cross‑attention/external‑KV conditioning interface, %making it a natural host for methods like OminiControl.
% In sum, U‑Net–style external condition branches (ControlNet/Adapter) and DiT/FLUX‑style token‑level condition injection (IC‑LoRA and OminiControl) advance from the angles of parameter‑efficient adaptation and representation unification, converging on the same goal: strengthening the conditional capacity of $\varepsilon_\theta(x_t,t\mid c)$
% while staying compatible with modern denoisers.

\section{Dataset}

In this study, we leverage three distinct EEG datasets tailored to specific visual synthesis objectives: image generation, image editing, and stylization. Specifically, we utilize the EEGCVPR40\cite{CVPR40,CVPR402} dataset for generation, the LoongX dataset\cite{loongx} for editing, and a newly curated dataset, \textit{Style-EEG}, for style transfer.

\textbf{EEGCVPR40.} The EEGCVPR40 dataset provides high-density, short-term neural recordings serving as the foundation for our image generation task. It comprises EEG signals from six subjects viewing images from 40 ImageNet\cite{imagenet} classes (50 images per class). Signals were acquired at a 1 kHz sampling rate using a 128-channel system, with a 0.5-second window recorded post-stimulus onset. Since the dataset typically features images with single dominant objects, it likely encourages subjects to process visual stimuli through semantic labeling and abstract representation, providing a strong basis for class-conditional generation.

\textbf{LoongX.} For the image editing task, we employ LoongX\cite{loongx}, a multimodal dataset containing 23,928 samples (22,728 training / 1,200 testing) sourced from 12 participants. The data integrates EEG signals with synchronized PPG\cite{ppg}, fNIRS\cite{fnir}, and motion data to capture comprehensive physiological cues during cognitive editing processes. The EEG component was recorded via four electrodes (Pz, Fp2, Fpz, Oz) at 250 Hz. Notably, data collection involved participants viewing edited images while concurrently reading editing instructions; this setup implies that the neural signals in LoongX intrinsically encode text-mediated intermediate representations.

\textbf{Style-EEG.} To bridge the gap in style-oriented neural decoding, we introduce the \textit{Style-EEG} dataset. We conducted a dedicated data acquisition experiment with five participants (mean age = 23.4 years), collecting neural responses to 1,000 groups of stylized images sourced from Midjourney\cite{midjourney}. Each group contained four images sharing a consistent artistic style. Stimuli were presented for 1 second, with the central 0.5-second segment extracted to ensure signal stability. The protocol included a 2-second inter-stimulus interval and periodic breaks to mitigate fatigue. In total, the dataset comprises 20,000 EEG segments, split into 15,000 for training and 5,000 for testing.

\textbf{Data Acquisition and Preprocessing.} All data collection procedures were approved by the institutional ethics committee, and participants provided informed consent. For \textit{Style-EEG}, signals were recorded using a 64-channel cap at 1 kHz, with five channels allocated for ECG and EOG monitoring. Preprocessing involved band-pass filtering (1--80 Hz) to remove drifts and high-frequency noise, alongside notch filtering (48--52 Hz) to eliminate power-line interference, ensuring high signal fidelity for downstream model training.
\section{Methodology}

\subsection{Preliminaries}
\textbf{Unified Conditional Generation--Editing.}
Given a natural-language instruction $I_{\text{txt}}$ and an optional context image $y$, our goal is to generate a target image $x$ by modeling $p_\theta(x \mid y, I_{\text{txt}})$, where $y \in \mathcal{X} \cup \{\varnothing\}$.
When $y\neq\varnothing$, the network performs image-guided editing; when $y=\varnothing$, it reduces to text-to-image generation.

\textbf{Latent Tokenization and 3D RoPE.}
Following FLUX, we operate in the latent space of a convolutional autoencoder $E$.
An image is encoded into a latent map $\mathbf{z}\in\mathbb{R}^{C\times H\times W}$, partitioned into $p\times p$ patches, and projected to a token sequence.
We use 3D rotary position embeddings (3D RoPE)~\cite{RoPE} with triplets $u=(b,h,w)$, where the \emph{block index} $b$ effectively separates target and context blocks:
\begin{equation}
u_x=(0,h,w), \qquad u_{y_i}=(i,h,w), \ \ i=1,\ldots,N.
\end{equation}
This formulation preserves the internal 2D structure of each block while enabling flexible multi-context conditioning.

\begin{figure*}[t]
  \centering
\includegraphics[width=1\textwidth,trim=0mm 26mm 0mm 32mm,clip]{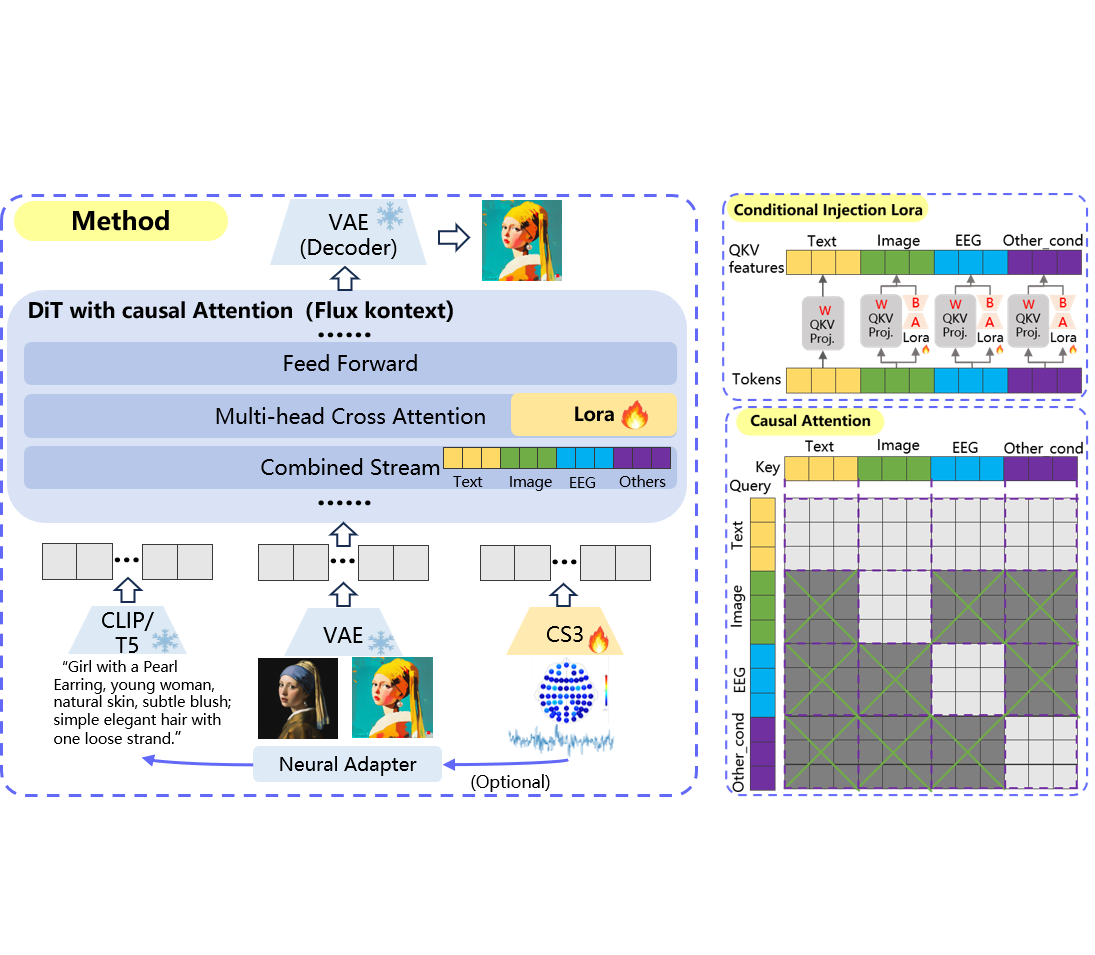}
  \caption{An overview of our proposed Unified Neural Signal-Guided Image Generation, Editing, and Stylization framework based on Diffusion Transformers, termed Uni-Neur2Img.}
  \label{fig:method}
\end{figure*}

\subsection{Uni-Neur2Img Framework}
The overall architecture is illustrated in Fig. \ref{fig:method}.

\textbf{Dual-text Encoder.}
We condition the FLUX.1-Kontext\cite{fluxkontext} backbone using a dual-encoder setup.
Given input text $I_{\text{txt}}$, T5\cite{t5} yields a token-level sequence $\mathbf{T}=\mathrm{T5}(I_{\text{txt}})\in\mathbb{R}^{M\times d}$, and CLIP\cite{clip} provides a pooled global vector $\mathbf{g}=\mathrm{CLIP}(I_{\text{txt}})\in\mathbb{R}^{d_g}$.
The full condition set is defined as:
\begin{equation}
\mathbf{c}=\{\mathbf{T},\mathbf{g},\mathbf{Z}_y,\mathbf{Z}_e\},
\end{equation}
where $\mathbf{Z}_y$ represents tokenized context-image latents and $\mathbf{Z}_e$ denotes optional EEG latents.

\textbf{Latent Path and Objective.}
We adopt the rectified flow objective~\cite{sd3,flux}.
Given the target image $x$, the encoder yields $\mathbf{z}_0=E(x)$.
We sample a time step $t \in [0,1]$ (mapped to noise scale $\sigma$) and construct a linear trajectory:
\begin{equation}\label{eq:latent_path}
\mathbf{z}_\sigma=(1-\sigma)\,\mathbf{z}_0+\sigma\,\boldsymbol{\epsilon}, \qquad \boldsymbol{\epsilon}\sim\mathcal{N}(0,I).
\end{equation}

The normalized step (either $t$ or $\sigma$) is fed to the backbone as in FLUX.

\textbf{Modality-aware Packing.}
Latents are flattened into sequences with image\_ids encoding $(h,w)$ positions and modality types.
EEG signals are processed by a lightweight CS3\cite{cs3} encoder to produce $\mathbf{Z}_e$, which is aligned with the image token layout but distinguished via index offsets.

\textbf{Efficient Fine-tuning.}
We inject LoRA adapters into the FLUX backbone's attention blocks.
We optimize only the LoRA parameters and the EEG encoder using AdamW, keeping the backbone frozen.

\textbf{Objective (flow matching).}
We adopt the rectified flow-style objective as in SD3/FLUX\cite{sd3}\cite{flux}:
the network $v_\theta$ predicts the velocity field on $\mathbf{z}_\sigma$
toward the noise endpoint. With a schedule-coupled weight $w(\sigma)$,
For completeness, we restate the flow-matching loss:
\begin{equation}
\mathcal{L}_\text{FM}
=\mathbb{E}_{\sigma,\boldsymbol{\epsilon}}
\Big[w(\sigma)\,\big\|v_\theta(\mathbf{z}_\sigma,\sigma,\mathbf{c})
-(\boldsymbol{\epsilon}-\mathbf{z}_0)\big\|_2^2\Big],
\end{equation}
where $\mathbf{z}_\sigma$ is defined in Eq.~\eqref{eq:latent_path},
$\mathbf{c}=\{\mathbf{T},\mathbf{g},\mathbf{Z}_y,\mathbf{Z}_e\}$,
and $v_\theta$ shares the same dimensionality as $\mathbf{z}_\sigma$
(either flattened to $d{=}C\!\times\!H\!\times\!W$ or kept in tensor form).

\subsection{Neural Adapter}
To enable generation from neural signals without explicit text, we propose a Neural Adapter that maps signals directly into the textual representation space.
Unlike GWIT\cite{GWIT} which decodes signals to discrete text, we align embedding spaces to bypass the error-prone text decoding step.

Formally, let $\mathbf{f}_{v} = \text{CLIP}(I_{\text{txt}})$ and $\mathbf{f}_{t} = \text{T5}(I_{\text{txt}})$ be the target textual features.
Given a raw neural signal sequence $\mathbf{N} \in \mathbb{R}^{T \times D_n}$, our adapter produces a unified representation:
\begin{equation}
\mathbf{f}_{\text{neural}} = \text{Adapter}\bigl(\text{NeuralDecoder}(\mathbf{N})\bigr).
\end{equation}
During neural-guided inference, we explicitly substitute the text features with projections of the neural features:
\begin{equation}
\mathbf{f}_{v} \leftarrow \text{Proj}_v(\mathbf{f}_{\text{neural}}), \quad \mathbf{f}_{t} \leftarrow \text{Proj}_t(\mathbf{f}_{\text{neural}}),
\end{equation}
where $\text{Proj}_{v/t}$ align dimensions to $d_g$ and $d$ respectively.

\subsection{Condition Injection via LoRA}

\textbf{Attention Formulation.}
Let the input sequence be $\mathbf{S}\in\mathbb{R}^{L\times d_\text{model}}$. The standard attention mechanism computes query, key, and value states via linear projections. For the $j$-th head ($j=1,\ldots,h$), these are defined as:
\begin{equation}\label{eq:qkv_defs}
Q^{(j)}=\mathbf{S}W_Q^{(j)},\quad
K^{(j)}=\mathbf{S}W_K^{(j)},\quad
V^{(j)}=\mathbf{S}W_V^{(j)},
\end{equation}
where $W_Q^{(j)}, W_K^{(j)} \in \mathbb{R}^{d_\text{model}\times d_k}$ and $W_V^{(j)} \in \mathbb{R}^{d_\text{model}\times d_v}$. The outputs are concatenated and projected by $W_O \in \mathbb{R}^{h d_v \times d_\text{model}}$ to form the final result.

\textbf{Low-Rank Adaptation.}
To inject conditional signals without disrupting the pre-trained weights, we freeze the original matrices $W_0 \in \{W_Q, W_K, W_V, W_O\}$ and introduce a parallel low-rank branch.
For any frozen weight matrix $W_0 \in \mathbb{R}^{d_\text{in} \times d_\text{out}}$, the adapted weight $W'$ is parameterized as:
\begin{equation}\label{eq:lora_update}
W' = W_0 + \frac{\alpha}{r} A B,
\end{equation}
where $A \in \mathbb{R}^{d_\text{in} \times r}$ and $B \in \mathbb{R}^{r \times d_\text{out}}$ are trainable matrices with rank $r \ll \min(d_\text{in}, d_\text{out})$.
By learning $A$ and $B$, the model effectively aligns the new modalities (e.g., EEG latents) into the requisite semantic space defined by the frozen $W_0$.

\subsection{Causal Mutual Attention}

Let the concatenated token sequence be
\begin{equation}
\mathbf{S}=\big[\mathbf{Z}_x;\mathbf{Z}_{y_1};\ldots;\mathbf{Z}_{y_N};\mathbf{Z}_e;\mathbf{T}\big],
\end{equation}
and denote block index sets by
$\mathcal{B}_x,\mathcal{B}_{y_i},\mathcal{B}_e,\mathcal{B}_{\text{txt}}$.
We introduce a structured mask to aggregate conditions via text tokens while isolating cross-condition interactions:
\begin{equation}\label{eq:mutual_mask}
M^{\text{mut}}_{ij}=
\begin{cases}
0, & i,j\in\mathcal{B}_b\ \text{for some block } b,\\
0, & i\in\mathcal{B}_{\text{txt}},\\
-\infty, & \text{otherwise}.
\end{cases}
\end{equation}
The attention is computed as
\begin{equation}\label{eq:attn}
\mathrm{Attn}(Q,K,V)=\mathrm{Softmax}\!\left(QK^\top/\sqrt{d_k}+M^{\text{mut}}\right) V.
\end{equation}
This blockwise mask balances conditional aggregation (through text tokens) and conditional isolation (by suppressing direct cross-condition attention), improving stability and efficiency during inference.

\section{Experiment}
\begin{table*}[htbp!]
  \centering
  \caption{Comparison on EEGCVPR40. Higher IS/ACC is better, lower FID/LPIPS is better. ``\textemdash'' means not reported.}
  \label{tab:eegcvpr40}

  % —— 仅作用于表格主体的局部分组 —— 
  \begingroup
  \setlength{\tabcolsep}{10pt}
  \renewcommand{\arraystretch}{1.12}
  % 固定表格文字为 10pt（行距可按需调，这里用 12pt）
  \fontsize{10pt}{12pt}\selectfont
  % 防止单元格或表注内出现 \scriptsize / \footnotesize 把字号缩小
  \let\scriptsize\normalsize
  \let\footnotesize\normalsize

  \begin{threeparttable}
  % 关键：仅缩放表格主体到 0.85× 宽度
  \resizebox{1\linewidth}{!}{%
  \begin{tabular}{llcccc}
    \toprule
    & \textbf{Model} & \textbf{IS} $\uparrow$ & \textbf{FID} $\downarrow$ & \textbf{ACC} $\uparrow$ & \textbf{LPIPS} $\downarrow$ \\
    \midrule
    \multirow{4}{*}{\makecell[l]{\textbf{EEGCVPR40}\\(Single)}}
      & DreamDiffusion\cite{dreamdiffusion}      & \textemdash & \textemdash & 0.45 & \textemdash \\
      & BrainVis\cite{brainvis}           & 31.52 & 121.02 & 0.49 & \textemdash \\
      & GWIT \cite{GWIT}          & 33.32  & 80.47  & 0.91 & 0.81 \\
      & \textbf{Uni-Neur2Img} & \textbf{35.19} (+5.61\%) & \textbf{74.70} (-7.17\%)  & \textbf{0.93} (+2.20\%)  & \textbf{0.75} (-7.41\%) \\
    \midrule
    \multirow{5}{*}{\makecell[l]{\textbf{EEGCVPR40}\\(Multi)}}
      & NeuroImagen \cite{NeuroImagen}        & 33.50 & \textemdash & 0.85       & \textemdash \\
      & EEGStyleGAN-ADA \cite{EEGStyleGAN-ADA}    & 10.82 & 174.13     & \textemdash & \textemdash \\
      & GWIT \cite{GWIT}                   & 33.87 & 78.11 & 0.91 & 0.77 \\
      & \textbf{Uni-Neur2Img} & \textbf{36.19} (+6.85\%) & \textbf{70.72} (-9.46\%)  & \textbf{0.92} (+1.10\%) & \textbf{0.73} (-5.2\%) \\
    \bottomrule
  \end{tabular}
  }% end of resizebox
   %\begin{tablenotes}
    %\item Higher IS/ACC is better; lower FID/LPIPS is better. ``\textemdash'' indicates the metric is not reported.
   %\end{tablenotes}
  \end{threeparttable}

  \endgroup
\end{table*}

\begin{figure*}[t]
  \centering
  \includegraphics[width=\textwidth]{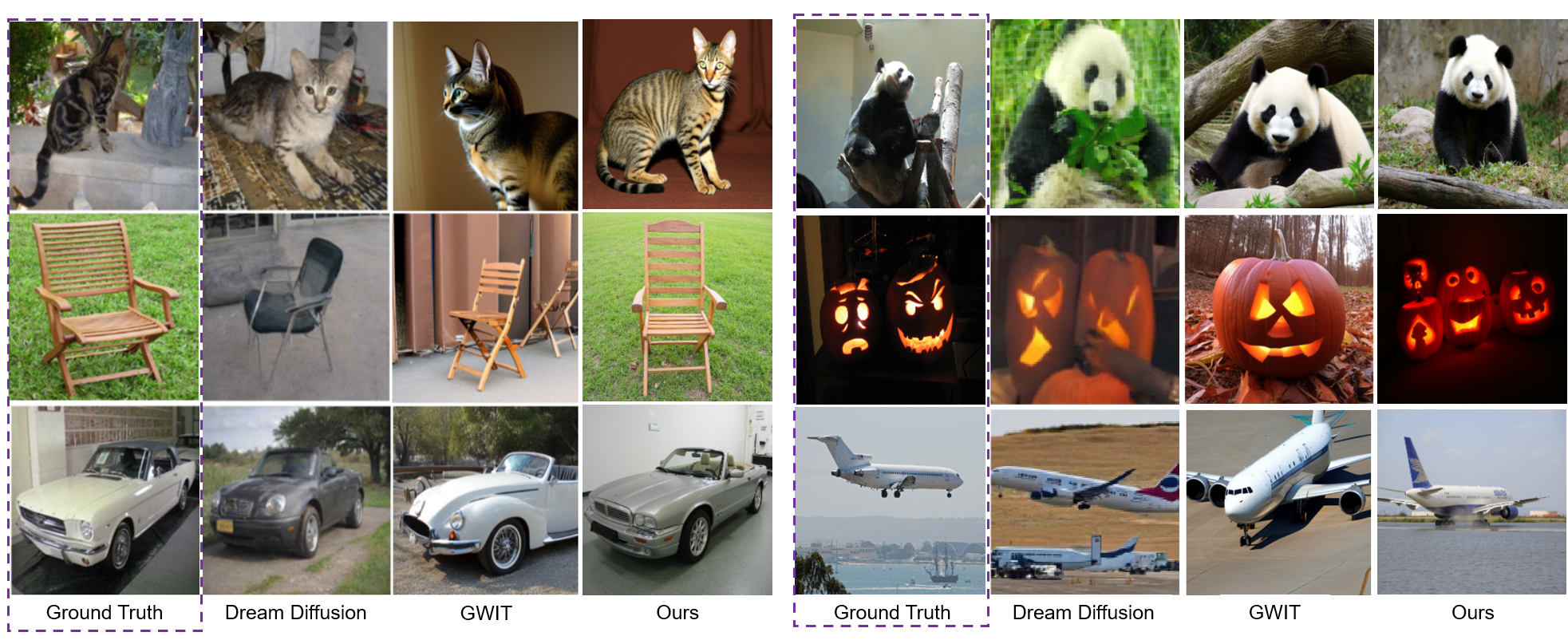} % 因为有 \graphicspath，无需写 figs/
  \caption{EEG-Driven Image Generation on the CVPR40 Dataset: Qualitative Comparison of Uni-Neur2Img with DreamDiffusion, GWIT, and Original Images.}
  \label{fig:exp1}
\end{figure*}
\subsection{Implementation Details}
We employ FLUX.1 dev and FLUX.1 kontext as the pre-trained Diffusion Transformer (DiT) and fine-tune it separately for each spatial or subject condition. Training is conducted on $8\times$ A100 (80\,GB) GPUs with an input resolution of $512\times512$ and a per-GPU batch size of 1 (effective batch size 4 without gradient accumulation). We use the AdamW optimizer with learning rate $1\times 10^{-4}$. We adopt bf16 mixed precision and LoRA adaptation with rank 128. Gradients are clipped at a norm of 1.0. The learning rate follows a constant schedule with a 500-step warm-up.

\begin{table*}[t]
  \centering
  \caption{Comparison between baseline methods and Uni-Neur2Img paradigms based on Loongx Dataset: (i)Text Only (During fine-tuning, EEG signals were set to all zeros.) (ii) Pure EEG and (iii) EEG signals enhanced by text.}
  \label{tab:loongx}
  \setlength{\tabcolsep}{10pt}
  \renewcommand{\arraystretch}{1.15}
  \begin{threeparttable}
  \resizebox{\textwidth}{!}{
  \begin{tabular}{lccccc}
    \toprule
    \textbf{Methods} & \textbf{L1} ($\downarrow$) & \textbf{L2} ($\downarrow$) & \textbf{CLIP-I} ($\uparrow$) & \textbf{DINO} ($\uparrow$) & \textbf{CLIP-T} ($\uparrow$) \\
    \midrule
    OminiControl (Text)   & 0.2632  & 0.1161 & 0.6558  & 0.4636  & 0.2549  \\
    \textbf{Uni-Neur2Img(Text Only)} & \textbf{0.1827} {\scriptsize (-30.59\%)} & \textbf{0.0690} {\scriptsize (-40.57\%)} & \textbf{0.7259} {\scriptsize (+10.69\%)} & \textbf{0.5673} {\scriptsize (+22.37\%)} & \textbf{0.2563} {\scriptsize (+0.55\%)} \\
    \midrule
    LoongX (Pure EEG)  & 0.2641 & 0.1078  & 0.5457  & 0.2963 & 0.2251  \\
    LoongX (Neural Signals) & 0.2509 & 0.1029 & 0.6605& 0.4812 & 0.2436  \\
    LoongX (Neural+Speech)  & 0.2594 & 0.1080  & 0.6374  & 0.4205 & \textbf{0.2588}  \\
     \textbf{Uni-Neur2Img (Pure EEG)} & \textbf{0.2122} {\scriptsize (-19.65\%)}& \textbf{0.0854} {\scriptsize (-20.78\%)} & \textbf{0.6475} {\scriptsize (+18.65\%)} & \textbf{0.4442} {\scriptsize (+44.91\%)} & \textbf{0.2280} {\scriptsize (-10.55\%)}\\
     \textbf{Uni-Neur2Img (EEG+Text)} & \textbf{0.1466} {\scriptsize (-44.49\%)}& \textbf{0.0551} {\scriptsize (-48.89\%)} & \textbf{0.7446} {\scriptsize (+36.44\%)} & \textbf{0.6020} {\scriptsize (+103.17\%)} & 
     0.2522 {\scriptsize (+12.03\%)}\\
    \bottomrule
  \end{tabular}}
  \end{threeparttable}
\end{table*}

\begin{figure*}[t]
  \centering
  % trim 的顺序是：left bottom right top；建议用 mm 为单位
  \includegraphics[width=\textwidth,trim=0mm 0mm 0mm 0mm,clip]{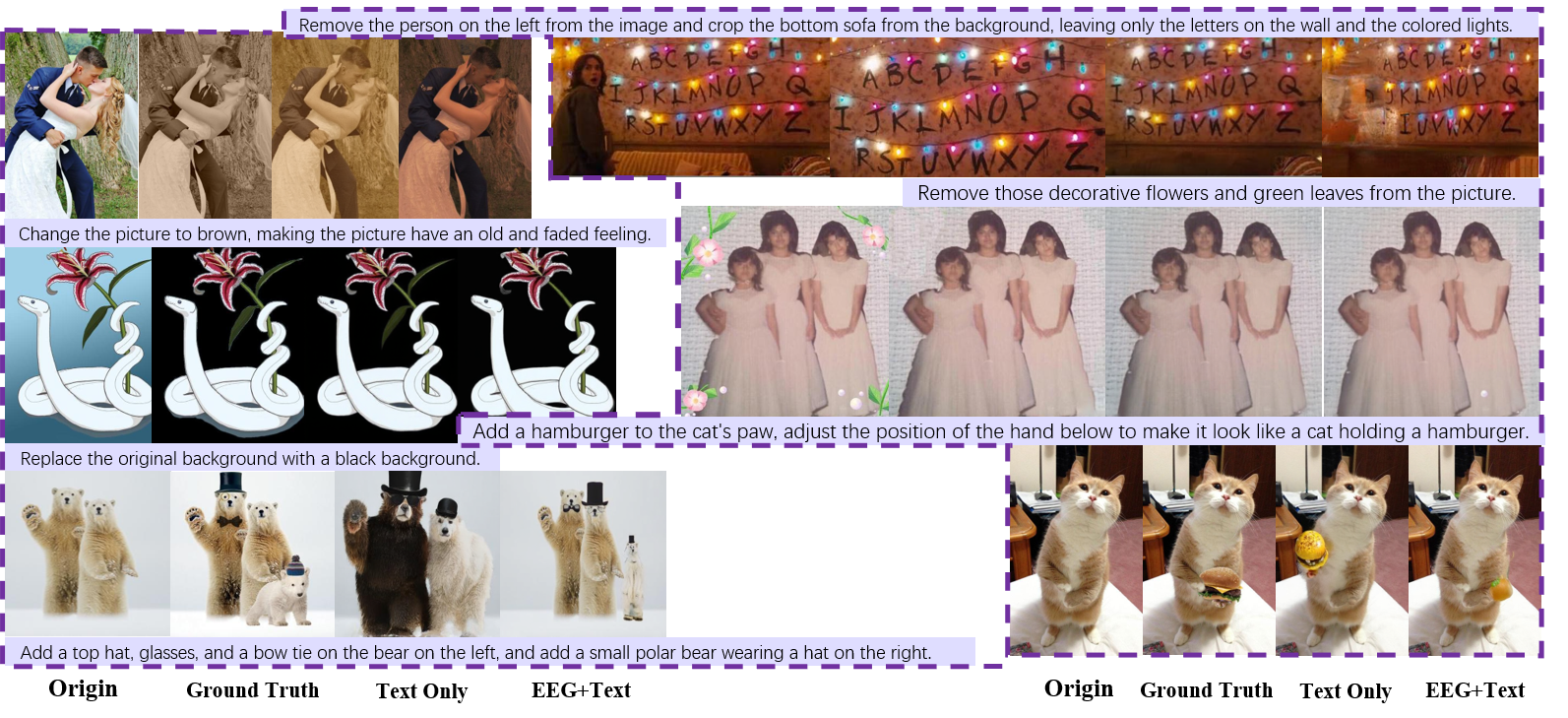}
  \caption{Qualitative Comparison of Uni-Neur2Img on the LoongX Dataset: Original, Ground Truth, EEG+Text, and EEG-Only Results Across (a) Background, (b) Object, (c) Global, and (d) Text Editing.}
  \label{fig:exp2}
\end{figure*}

\subsection{EEG Guided Image Generation}
% \textbf{Method.}
% We follow the pipeline described in the \emph{Methods} section and study Flow-Matching EEG-conditioned image synthesis on top of a pretrained VAE and \textsc{FLUX} Transformer.
% Given paired samples \(\{e,x\}\), where \(e \in \mathbb{R}^{C \times L}\) denotes an EEG signal with \(C\) channels and \(L\) time steps, and \(x \in \mathbb{R}^{H_x \times W_x \times 3}\) is the corresponding RGB image, we first encode \(x\) with VAE into a latent \(z\) downsampled by \(16\times\) and rearrange it into the token layout required by \textsc{FLUX}.
% For the EEG condition \(e\), a CS3 encoder maps it to a conditional feature map that is resolution-aligned with the image latent; we further construct 2D positional indices for both the main and conditional streams to ensure spatial alignment.
% In addition, we introduce a coarse-grained control \(c_\ell\) following \cite{eeglabel}, realized as a caption of the form ``Image of [label]''.
% The label is predicted by a pretrained (frozen) EEG-to-image decoder, and the resulting text is fed into the text branch (Fig.~\ref{fig:method}).
% To preserve the base-model capacity while minimizing trainable overhead, we insert multi-branch \textsc{LoRA} adapters only into the attention \(Q/K/V/\mathrm{Proj}\) paths (covering both double-stream and single-stream blocks), keep all remaining parameters frozen, and update only the \textsc{LoRA} modules together with the CS3 encoder, enabling efficient cross-modal conditioning.

\textbf{Quantitative and Qualitative Comparison.} 
 Table \ref{tab:eegcvpr40} reports quantitative comparisons on the EEGCVPR40 benchmark under both single‑subject and multi‑subject settings. Across all reported metrics, our Uni‑Neur2Img consistently outperforms prior EEG‑to‑image approaches, including DreamDiffusion\cite{dreamdiffusion}, BrainVis\cite{brainvis}, GWIT\cite{GWIT}, NeuroImagen\cite{NeuroImagen} and EEGStyleGAN‑ADA\cite{EEGStyleGAN-ADA}. On EEGCVPR40 (Single), Uni‑Neur2Img achieves the best performance. %Inception Score, surpassing the strongest baseline GWIT by +5.61\%, while reducing FID by 7.17\% and improving classification accuracy by +2.20\%%. Compared with earlier methods such as BrainVis, the gains are even more pronounced, particularly in FID and ACC.
On EEGCVPR40 (Multi), the performance gap further widens: \text{Uni‑Neur2Img} achieves an IS improvement of 6.85\%, a FID reduction of 9.46\%,  an ACC gain of 1.10\%, and a LPIPS reduction of 5.2\% over GWIT, while also clearly outperforming NeuroImagen and others. 
Qualitative examples in Fig. \ref{fig:exp1} corroborate these quantitative trends. Images generated by Uni‑Neur2Img more faithfully preserve object categories and fine‑grained attributes such as cat pose, chair geometry, and car shape, producing sharper edges and cleaner backgrounds. In contrast, DreamDiffusion and GWIT often yield blurrier textures, shape distortions (e.g., deformed chairs and pumpkins), or mismatched details, highlighting the advantage of our text‑and‑neural joint representation for high‑fidelity EEG‑conditioned image synthesis.

\subsection{Neural Guided Image Editing}
% \textbf{Method.} Compared to pure image synthesis, our image editing pipeline differs in three aspects. First, the text prompt is provided directly by the dataset and is encoded into conditional features \( \mathbf{c}_{\text{text}} \) that are spatially aligned to the latent image resolution. Second, the visual input is organized as a pair of \emph{original} and \emph{ground-truth} (used as the target/inference reference) images, both mapped by a shared VAE encoder \(E_{\phi}\) into the same latent space to obtain \( \mathbf{z}_{\text{ori}} \) and \( \mathbf{z}_{\text{gt}} \), enabling content preservation and target supervision. Finally, for the LoongX dataset with multimodal neural signals—EEG, fNIRS, PPG, and motion physiological signals—we employ a CS3-based encoder to represent all neural streams \( \mathbf{c}_{\text{neuro}} \), putting them into other\_condition branch and conditioning embeddings into a unified latent representation that remains consistent with the text embedding. The resulting representations from all branches are then injected into a FLUX.1-Kontext backbone to perform conditioned editing.

\textbf{Quantitative and Qualitative Comparison.}  As shown in Table \ref{tab:loongx}, when compared to the baselines OmniControl and LoongX, our Uni‐Neur2Img consistently exhibits superior performance under both pure neural and \text{speech‐enhanced} settings. In the Pure EEG scenario, \text{Uni‑Neur2Img} reduces the L1/L2  error by about 19.65\%/20.78\% over LoongX, while improving CLIP‑I and DINO by 18.65\% and 44.91\%, respectively, with comparable CLIP‑T, indicating that our decoder can more effectively exploit EEG signals for both pixel‑level accuracy and perceptual quality. When further incorporating speech, the gains become even more pronounced: compared to LoongX (Neural+Speech), Uni‑Neur2Img (Neural+Text) yields additional 43.32\%/35.4\% reductions in L1/L2 and 16.8\%/43.1\% improvements in CLIP‑I/DINO, with only a minor 2.55\% drop in CLIP‑T, demonstrating that our cross‑modal fusion strategy can effectively leverage speech to enhance neural representations without the degradation in pixel and image‑level semantics observed in LoongX.
Moreover, relative to the strong text-only upper-bound baseline Uni-Neur2Img (Text Only), Uni-Neur2Img (Neural+Text) yields additional L1/L2 reductions of 19.75\% and 20.14\%, despite a 1.5\% decrease in CLIP-T, indicating that our method surpasses text-level semantic fidelity when incorporating EEG signals and even exceeds text-driven models in generation accuracy. Overall, as shown in Fig. \ref{fig:exp2} these results underscore that Uni-Neur2Img provides a more accurate and robust solution for neural-to-image generation under both EEG-only and speech-enhanced neural settings compared to LoongX.

\begin{table*}[t]
  \centering
  \caption{Quantitative Evaluation of EEG-based Image stylization (mean ± std; ↑ higher is better, ↓ lower is better).}
  \label{tab:style}
  \setlength{\tabcolsep}{5pt}
  \renewcommand{\arraystretch}{1.15}
  \begin{threeparttable}
  \resizebox{0.9\linewidth}{!}{%
  %{\fontsize{10}{12}\selectfont
    \begin{tabular*}{\textwidth}{@{\extracolsep{\fill}}lccccc}
      \toprule
      \textbf{Methods} & \textbf{CLIP-I} ($\uparrow$) & \textbf{DINO} ($\uparrow$) & \textbf{PSNR} ($\uparrow$) & \textbf{SSIM} ($\uparrow$) & \textbf{PDist} ($\downarrow$) \\
      \midrule
      \textbf{Uni-Neur2Img (Ours)} & 0.7487 & 0.4089 & 9.4166 & 0.2573 & 0.7332 \\
      \bottomrule
    \end{tabular*}
  }
  \end{threeparttable}
\end{table*}

\begin{figure*}[t]
  \centering
  % trim 的顺序是：left bottom right top；建议用 mm 为单位
  \includegraphics[width=\textwidth,trim=0mm 0mm 0mm 0mm,clip]{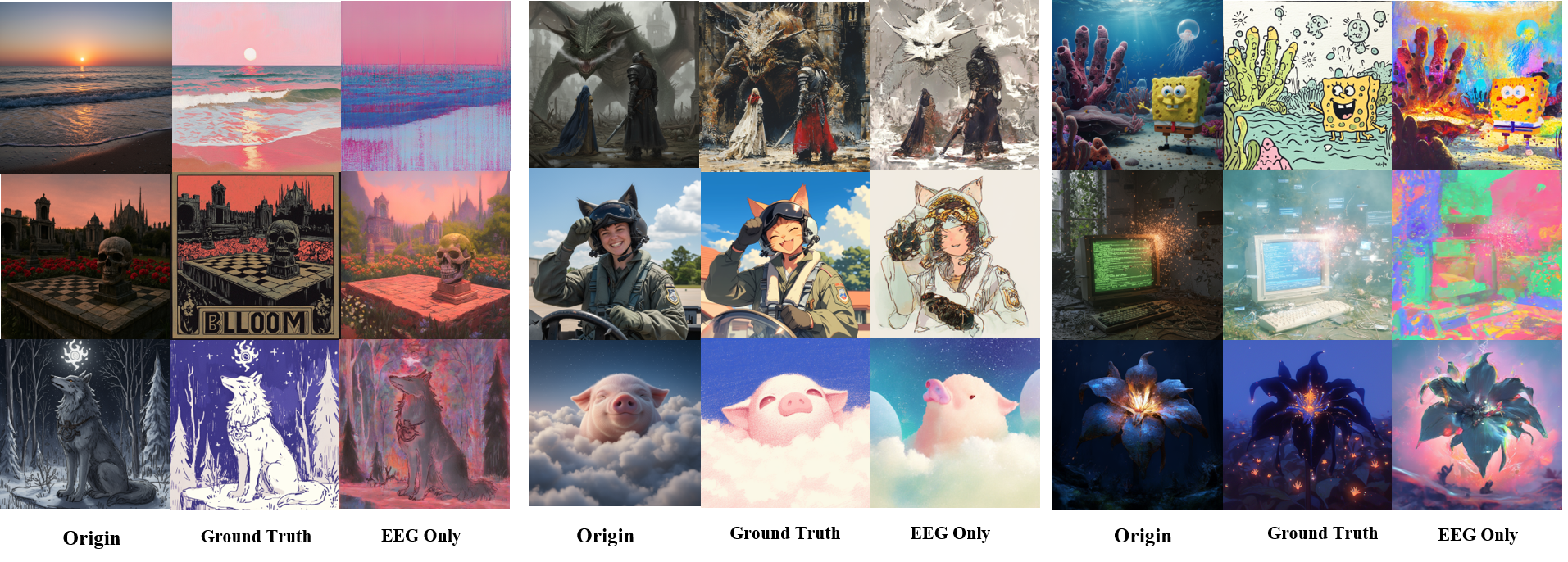}
  \caption{Comparison of EEG-Based Style Transfer Results with Ground Truth Stylizations and Original Images on the EEG-Style Dataset.}
  \label{fig:style}
\end{figure*}
\subsection{EEG Guided Image Stylization}
\textbf{Quantitative and Qualitative Comparison.}
With EEG recorded while subjects viewed style images, our EEG‑only pipeline Uni‑Neur2Img (EEG Only) results show in Table \ref{tab:style}. Although pixel‑level fidelity (PSNR/SSIM) is low-consistent with the fact that stylization does not require pixel alignment—the method preserves strong alignment with the target style in semantic/perceptual space (CLIP‑I:0.7487 ± 0.0865, DINO:0.4089 ± 0.1576, PDist: 0.7332 ± 0.1082), and the standard deviations reveal non‑trivial variability across styles. Qualitatively, the EEG generation reliably recover global color palettes, material appearance, and coarse composition, exhibiting consistent style biases across diverse content types (landscape, portrait, still life). In contrast, fine‑grained stylistic morphemes—sharp line art, repetitive patterns, and high‑frequency strokes—remain challenging, leading to occasional boundary distortions and texture over‑smoothing. Overall, as shown in Fig. \ref{fig:style}, the results verify that EEG signals contain decodable style representations and that perceptually plausible stylizations can be generated from EEG alone, while restoring intricate details and iconic motifs is the primary bottleneck.

\section{Ablation Study}

\subsection{EEG Guided Image Generation}
Complementing the Neural Adapter-based experiments in Table \ref{tab:eegcvpr40}, we further evaluate a challenging setting where \textbf{raw EEG signals} serve as the sole input to the Flux1.Kontext backbone for image generation. Table \ref{tab:cvpr402} presents the quantitative comparison on the CVPR40 multi-subject benchmark under this EEG-only configuration. 
Existing methods, such as Brain2Image-VAE/GAN, NeuroVision, Improved-SNGAN, and DCLS-GAN, struggle in this setting, yielding Inception Scores (IS) below 7 and lacking reported FID metrics. While the recent EEGStyleGAN-ADA baseline improves the IS to 10.82, it still suffers from significant distribution divergence (FID: 174.13). In contrast, our Uni-Neur2Img achieves a remarkable IS of 24.46 (+126.06\%) and a reduced FID of 148.87 (-14.51\%). These results demonstrate that our framework can generate images that are not only more realistic but also exhibit superior consistency with the underlying raw neural signals.
% 请确保在文档最前面（\documentclass 之后）加上：\usepackage{array}

\begin{table}[!ht]
  \centering
  % 保证在双栏中仅占一栏宽度
  \caption{EEG-only (no textual input) results on the CVPR40 multi-subject benchmark. ``\textemdash'' means not reported.}
  \label{tab:cvpr402}
  \resizebox{\columnwidth}{!}{%
  \setlength{\tabcolsep}{8pt}
  \renewcommand{\arraystretch}{1.15}
  % 【修改点1】这里改为 4 列：l (空列), m (方法名), c (IS), c (FID)
  % 如果想要第二列文字居中，可以将 m{3.6cm} 改为 >{\centering\arraybackslash}m{3.6cm}
  \begin{tabular}{l m{3.6cm} c c }
    \toprule
    % 第一列为空，为左侧竖排数据集留位
    & \textbf{Method} & \textbf{IS $\uparrow$} & \textbf{FID $\downarrow$}  \\
    \midrule
    % 左侧竖排数据集标签，跨越下面 6 行（根据行数修改 6）
    \multirow{7}{*}{\rotatebox[origin=c]{90}{\footnotesize \textbf{EEGCVPR40}}} 
      & \textbf{Brain2Image\texttt{-}VAE} \cite{B2Ivae}        & 4.49  & ---    \\
      & \textbf{Brain2Image\texttt{-}GAN } \cite{B2Igan,B2Ivae}    & 5.07      & ---    \\
      & \textbf{NeuroVision} \cite{Neurovis}                      & 5.15    & ---    \\
      & \textbf{Improved\texttt{-}SNGAN} \cite{sngan}         & 5.53     & ---    \\
      & \textbf{DCLS\texttt{-}GAN} \cite{dcls-gan}                 & 6.64    & ---    \\
      & \textbf{EEGStyleGAN\texttt{-}ADA} \cite{EEGStyleGAN-ADA}       & 10.82 & 174.13  \\
      & \textbf{Uni-Neur2Image(Ours)}
      % 【修改点2】修复了这里的负号，并修正了 & 分隔符以匹配4列
      & \shortstack{\textbf{24.46}\\{\scriptsize (+126.06\%)}}
      & \shortstack{\textbf{148.87}\\{\scriptsize (-14.51\%)}}
    \\
    \bottomrule
  \end{tabular}%
  } % end resizebox 
\end{table}

\subsection{Neural Guided Image Editing}
\begin{table*}[t]
  \centering
  \caption{Comparison of Uni-Neur2Img paradigms based on Loongx Dataset with different input: (i)Text Only (During fine-tuning, EEG signals were set to all zeros.) (ii) Pure EEG (iii) EEG+fNIR and (iv) EEG signals enhanced by text.}
  \label{tab:loongx2}
  \setlength{\tabcolsep}{10pt}
  \renewcommand{\arraystretch}{1.15}
  \begin{threeparttable}
  \resizebox{\textwidth}{!}{
  \begin{tabular}{lccccc}
    \toprule
    \textbf{Methods} & \textbf{L1} ($\downarrow$) & \textbf{L2} ($\downarrow$) & \textbf{CLIP-I} ($\uparrow$) & \textbf{DINO} ($\uparrow$) & \textbf{CLIP-T} ($\uparrow$) \\
    \midrule
     \textbf{Uni-Neur2Img (EEG Only)} & 0.2122& 0.0854 & 0.6475 &0.4442 & 0.2280\\
     
     \textbf{Uni-Neur2Img (Text Only)} & \textbf{0.1827} {\scriptsize (-6.74\%)}& \textbf{0.0690} {\scriptsize (-4.80\%)} & \textbf{0.7259} {\scriptsize (+6.22\%)} & \textbf{0.5673} {\scriptsize (+18.52\%)} & 
     \textbf{0.2563} {\scriptsize (+1.80\%)}\\
     
      \textbf{Uni-Neur2Img (EEG+fNIR)} & \textbf{0.1979} {\scriptsize (-6.74\%)}& \textbf{0.0813} {\scriptsize (-4.80\%)} & \textbf{0.6878} {\scriptsize (+6.22\%)} & \textbf{0.5260} {\scriptsize (+18.52\%)} & 
     0.2321 {\scriptsize (+1.80\%)}\\
     
     \textbf{Uni-Neur2Img (EEG+Text)} & \textbf{0.1466} {\scriptsize (-30.91\%)}& \textbf{0.0551} {\scriptsize (-35.48\%)} & \textbf{0.7446} {\scriptsize (+15.00\%)} & \textbf{0.6020} {\scriptsize (+35.52\%)} & 
     0.2522 {\scriptsize (+10.61\%)}\\
    \bottomrule
  \end{tabular}}
  \end{threeparttable}
\end{table*}

\begin{figure*}[t]
  \centering
  % trim 的顺序是：left bottom right top；建议用 mm 为单位
  \includegraphics[width=\textwidth,trim=0mm 0mm 0mm 0mm,clip]{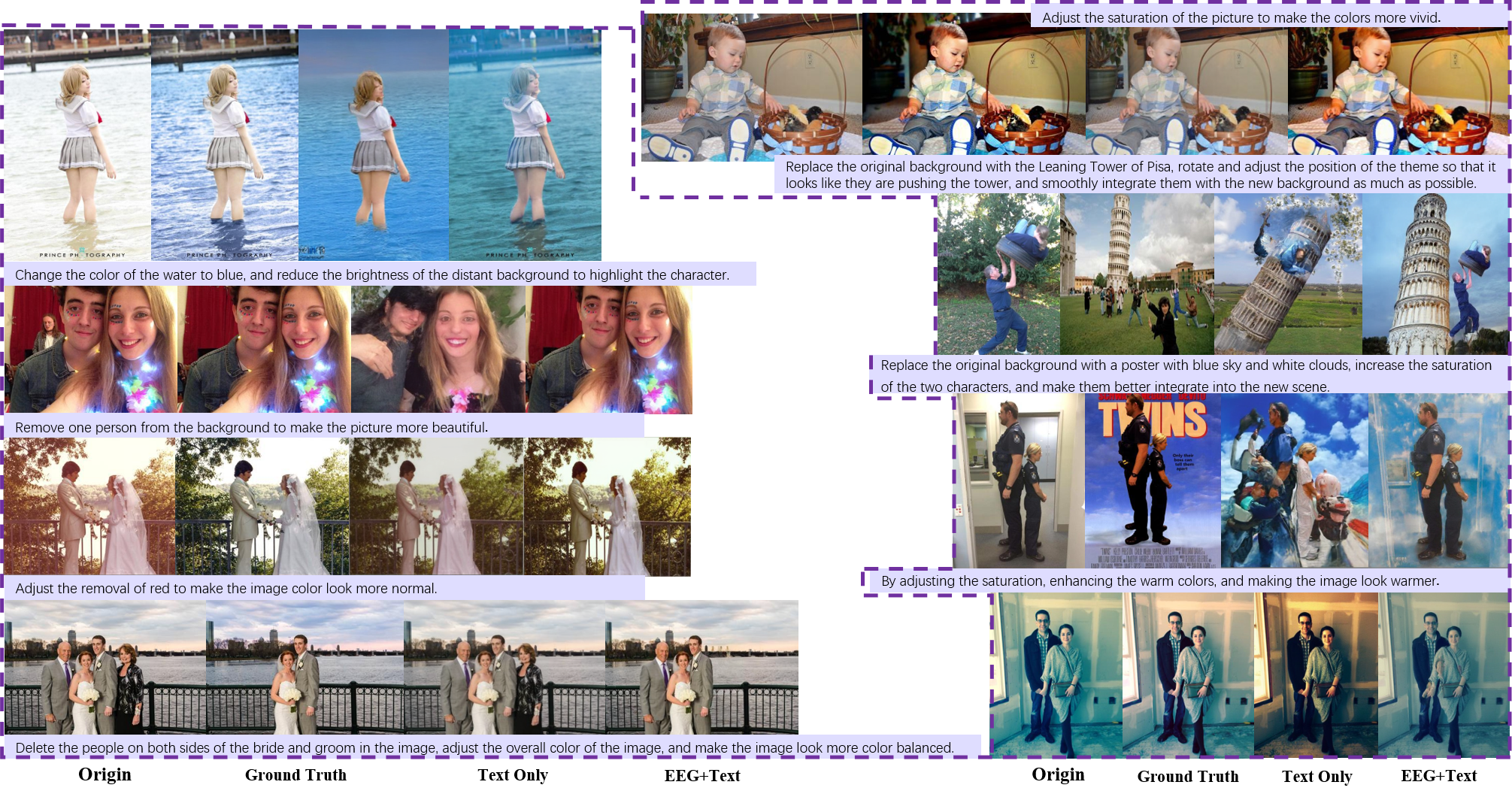}
  \caption{Qualitative Comparison of Uni-Neur2Img on the LoongX Dataset: Original, Ground Truth, EEG+Text, and EEG-Only Results Across (a) Background, (b) Object, (c) Global, and (d) Text Editing.}
  \label{fig:loongx2.1}
\end{figure*}

\textbf{Quantitative Analysis.}
We evaluate three variants of Uni-Neur2Img on the LoongX dataset: \textit{Pure EEG}, \textit{EEG+fNIR}, and \textit{Text-enhanced EEG+Text}, with detailed results summarized in Table \ref{tab:loongx2}. 
Following the findings of LoongX\cite{loongx}, which suggest that combining EEG and fNIR approximates the performance of full-modality fusion, we adopt EEG+fNIR as our multimodal neural input. Specifically, fNIR signals are injected via an auxiliary conditioning branch and fused with the EEG stream within the DiT backbone. 

As shown in Table \ref{tab:loongx2}, the introduction of fNIR signals improves reconstruction quality, reducing L1 and L2 errors by 6.74\% and 4.80\% respectively compared to the Pure EEG baseline, while boosting high-level semantic consistency (e.g., DINO score increases by 18.52\%).
Building on this, the \textit{EEG+Text} variant yields the most substantial gains. It reduces L1 and L2 errors by 30.91\% and 35.48\%, indicating significantly higher pixel-level fidelity. Meanwhile, semantic metrics see a dramatic surge, with CLIP-I, DINO, and CLIP-T improving by 15.00\%, 35.52\%, and 10.61\%, respectively. These results confirm that textual prompts provide critical semantic guidance for EEG representations, jointly enhancing both low-level reconstruction and high-level alignment.

\textbf{Qualitative Analysis.}
Figure \ref{fig:loongx2.1} presents additional qualitative comparisons between our \textit{Text-Only} baseline and the \textit{EEG+Text} setting, alongside original inputs and ground truth.
Across diverse editing scenarios, the superiority of the EEG+Text approach is evident.
\begin{itemize}
    \item \textbf{Low-level Color Editing:} For instructions such as ``change the water to blue'' or ``enhance saturation,'' the Text-Only baseline often produces incomplete modifications or global color deviations. In contrast, Uni-Neur2Img (EEG+Text) achieves precise local adjustments (e.g., changing water color while suppressing background brightness), closely matching the ground truth.
    \item \textbf{Structural Editing:} In tasks requiring object manipulation, such as ``replace the background with the Leaning Tower of Pisa,'' the Text-Only model tends to introduce artifacts or distort object boundaries. Our method effectively preserves the identity and pose of the main subjects while seamlessly compositing them into new scenes.
    \item \textbf{Global Stylization:} For global adjustments like ``make the image look warmer,'' our model produces results where color temperature and contrast strictly follow the user's intent without blurring local details.
\end{itemize}
Overall, these examples demonstrate that incorporating neural signals ensures image edits more faithfully reflect user intent while maintaining structural and semantic integrity.

\subsection{EEG Guided Image Stylization}

\begin{table*}[t]
  \centering
  \caption{Experimental results of Uni-Neur2Img with different input modalities on the EEG-Style single-subject dataset: (i)EEG only  (ii) Text Only (iii) EEG signals enhanced by text.}
  \label{tab:style2}
  \setlength{\tabcolsep}{10pt}
  \renewcommand{\arraystretch}{1.15}
  \begin{threeparttable}
  \resizebox{\textwidth}{!}{
  \begin{tabular}{lccccc}
    \toprule
    \textbf{Methods} & \textbf{CLIP-I} ($\uparrow$) & \textbf{DINO} ($\uparrow$) & \textbf{PSNR} ($\uparrow$) & \textbf{SSIM} ($\uparrow$) & \textbf{PDist} ($\downarrow$) \\
    \midrule
   \textbf{Uni-Neur2Img (EEG Only)} & 0.7487 & 0.4089 & 9.4166 & 0.2573 & 0.7332 \\
    \textbf{Uni-Neur2Img(Text Only)} & \textbf{0.8260} {\scriptsize (+9.36\%)} & \textbf{0.6179} {\scriptsize (+51.11\%)} & \textbf{9.2378} {\scriptsize (+1.90\%)} & \textbf{0.2733} {\scriptsize (+6.22\%)} & \textbf{0.7089} {\scriptsize (-3.31\%)} \\
    \textbf{Uni-Neur2Img(Text+EEG)} & \textbf{0.8182} {\scriptsize (-+9.28\%)} & \textbf{0.5741} {\scriptsize (+40.40\%)} & \textbf{11.0139} {\scriptsize (+16.96\%)} & \textbf{0.3168} {\scriptsize (+23.12\%)} & \textbf{0.6250} {\scriptsize (-14.76\%)} \\
    
    \bottomrule
  \end{tabular}}
  \end{threeparttable}
\end{table*}

\begin{table*}[t]
  \centering
  \caption{Experimental results of Uni-Neur2Img with different input modalities on the EEG-Style multi-subject dataset: (i)EEG only  (ii) Text Only (iii) EEG signals enhanced by text.}
  \label{tab:style-multi}
  \setlength{\tabcolsep}{10pt}
  \renewcommand{\arraystretch}{1.15}
  \begin{threeparttable}
  \resizebox{\textwidth}{!}{
  \begin{tabular}{lccccc}
    \toprule
    \textbf{Methods} & \textbf{CLIP-I} ($\uparrow$) & \textbf{DINO} ($\uparrow$) & \textbf{PSNR} ($\uparrow$) & \textbf{SSIM} ($\uparrow$) & \textbf{PDist} ($\downarrow$) \\
    \midrule
    \textbf{Uni-Neur2Img(Text Only)} & 0.8260 
    & \textbf{0.6179} & 9.2378  & 0.2733  & 0.7089 \\
    
   \textbf{Uni-Neur2Img(EEG Only-multi)} & 0.7975 & 0.5284  & \textbf{9.9694} {\scriptsize (+7.92\%)} & \textbf{0.2886}  {\scriptsize (+5.60\%)}& \textbf{0.6660} {\scriptsize (-6.05\%)} 
   \\
    
    \textbf{Uni-Neur2Img(Text+EEG-multi)} & \textbf{0.8347} {\scriptsize (+1.05\%)}
    &0.6102 %{\scriptsize (+40.40\%)} 
    & \textbf{11.0748} {\scriptsize (+19.89\%)} 
    & \textbf{0.3316} {\scriptsize (+21.33\%)}
    & \textbf{0.6069} {\scriptsize (-14.39\%)} 
    \\
    
    \bottomrule
  \end{tabular}}
  \end{threeparttable}
\end{table*}

\begin{figure*}[t]
  \centering
  % trim 的顺序是：left bottom right top；建议用 mm 为单位
  \includegraphics[width=\textwidth,trim=0mm 0mm 0mm 0mm,clip]{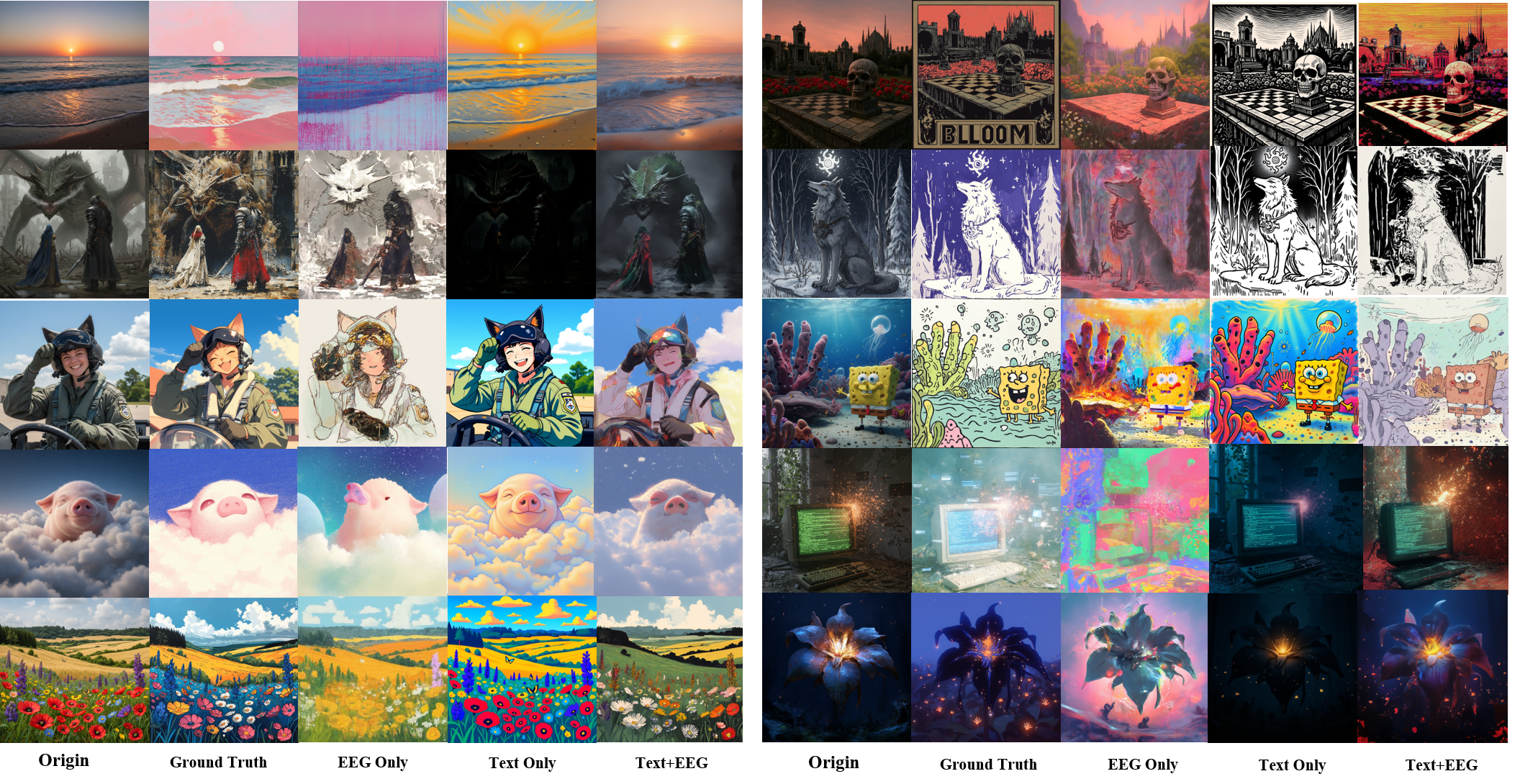}
  \caption{Qualitative results of Uni‑Neur2Img on the EEG‑Style dataset, comparing the original image, the ground‑truth target, and images generated under three input conditions: EEG Only, Text Only, and joint Text+EEG.}
  \label{fig:s2.1}
\end{figure*}

\textbf{Quantitative Analysis.} 
As presented in Table \ref{tab:style2}, incorporating textual input yields substantial improvements over the EEG-only setting across most metrics. While the \textit{Text Only} variant enhances global semantic consistency, the \textit{Text+EEG} configuration effectively synergizes the complementary strengths of both modalities. Specifically, Text+EEG maintains high semantic alignment (CLIP-I: 0.8182, +9.28\%; DINO: 0.5741, +40.40\%) while significantly boosting low-level fidelity. Notably, PSNR increases to 11.0139 (+16.96\%), SSIM rises to 0.3168 (+23.12\%), and PDist is markedly reduced to 0.6250 (-14.76\%). These metrics demonstrate that EEG signals and text are highly complementary: EEG signals are particularly effective at refining fine-grained structures and perceptual quality when guided by textual semantics.

We further extend our evaluation to the multi-subject scenario on the EEG-Style dataset (Table \ref{tab:style-multi}). A compelling observation is that the \textit{EEG Only-multi} variant already surpasses the text-driven baseline in reconstruction fidelity. Compared to \textit{Text Only}, \textit{EEG Only-multi} improves PSNR and SSIM by 7.92\% and 5.60\% respectively, and reduces PDist by 6.05\%. Although it shows a mild degradation in global semantic alignment (lower CLIP-I/DINO), this gap is relatively narrow, implying that EEG signals intrinsically encode rich stylistic priors. 
When fusing both modalities (\textit{Text+EEG-multi}), the model achieves superior performance across the board: CLIP-I is boosted to 0.8347, while PSNR and SSIM improve by 19.89\% and 21.33\% respectively. 
Crucially, these results suggest that EEG captures implicit stylistic attributes—such as structural layout, color composition, and texture statistics—that are inherently difficult to articulate linguistically. The fusion model thus benefits from a hierarchical interaction: text provides high-level semantic constraints, while EEG fills in the mid-to-low-level stylistic nuances.

\textbf{Qualitative Analysis.}
As shown in Fig. \ref{fig:s2.1}, visual comparisons further corroborate these quantitative findings. 
In the \textit{EEG Only} setting, generated images capture the general stylistic atmosphere but often suffer from high-frequency degradation, manifesting as blurry details or distorted structures. 
Conversely, the \textit{Text Only} baseline ensures semantic coherence (e.g., correct object categories) yet frequently misses subtle artistic textures or deviates from the specific ground-truth style. 
The \textit{Text+EEG} results achieve the optimal trade-off: sunset scenes exhibit sharper horizons and faithful color transitions, while complex scenes preserve both semantic correctness and stylistic fidelity. This confirms that incorporating neural signals allows Uni-Neur2Img to bridge the gap between abstract semantic intent and concrete perceptual appearance.

\section{Conclusion}

\textbf{Conclusion.}
In summary, our main contributions are three-fold:

1) We propose uni-Neur2Img, a unified neural framework that enables image generation, image editing, and image style transfer directly conditioned on neural signals. The framework achieves state-of-the-art performance on all three tasks, demonstrating its effectiveness and its potential to serve as a general backbone for these three tasks and potentially more neural-to-image applications.

2) We introduce, for the first time, a neural-conditioned image stylization task and construct EEG-Style, a dedicated dataset that provides a standardized benchmark for this setting.

3) Through experiments on image editing and styliztion, we show that EEG signals encode mid- and low-level visual information, such as structural layout, color composition, and texture distribution, more effectively than textual descriptions, especially for properties that are inherently hard to specify in language.

\textbf{Limitations and Future Work.}
A limitation of our study is that, due to equipment constraints, we were restricted to EEG signals rather than multimodal data (e.g., fNIRS). Future work should explore generation from fused neural modalities to further enhance decoding robustness. Moreover, investigating the explicit mapping between neural signals and specific interpretable image features remains a promising avenue for decoding the "black box" of visual perception.

{
    \small
    \bibliographystyle{ieeenat_fullname}
    \bibliography{main}
}
 \clearpage
\setcounter{page}{1}
\maketitlesupplementary

\section{EEG-Style—EEG Dataset Acquisition}
\label{sec:styledata}
To probe how much visual color and style information is carried by EEG, we collected an EEG‑Style sub‑corpus paired with stylized images.

\textbf{Participants \& Ethics.} Five healthy graduate student volunteers (2 female, 3 male; mean age $23.5\pm2.5$ years) took part in the study. All participants self-reported to be in good physical and mental health. Before the recording session, the full experimental procedure was explained, written informed consent was obtained, and each participant received monetary compensation for their time.
In this work, the attention level is quantified from EEG using a classical spectral index, defined as the ratio between the alpha-band (8–12 Hz) power and the theta-band (4–8 Hz) power. This index has been extensively adopted in cognitive neuroscience as a neural marker of attentional control and mental workload. For example, Raufi et al. \cite{atten} showed that both alpha/theta and theta/alpha power ratios correlate well with self-reported mental workload in EEG-based experiments. The resulting attention scores for each participant, together with their neural-signal-based image editing errors, are summarized in Table \ref{tab:subjects}.

\begin{figure}[!ht]
  \centering
  \includegraphics[width=\columnwidth]{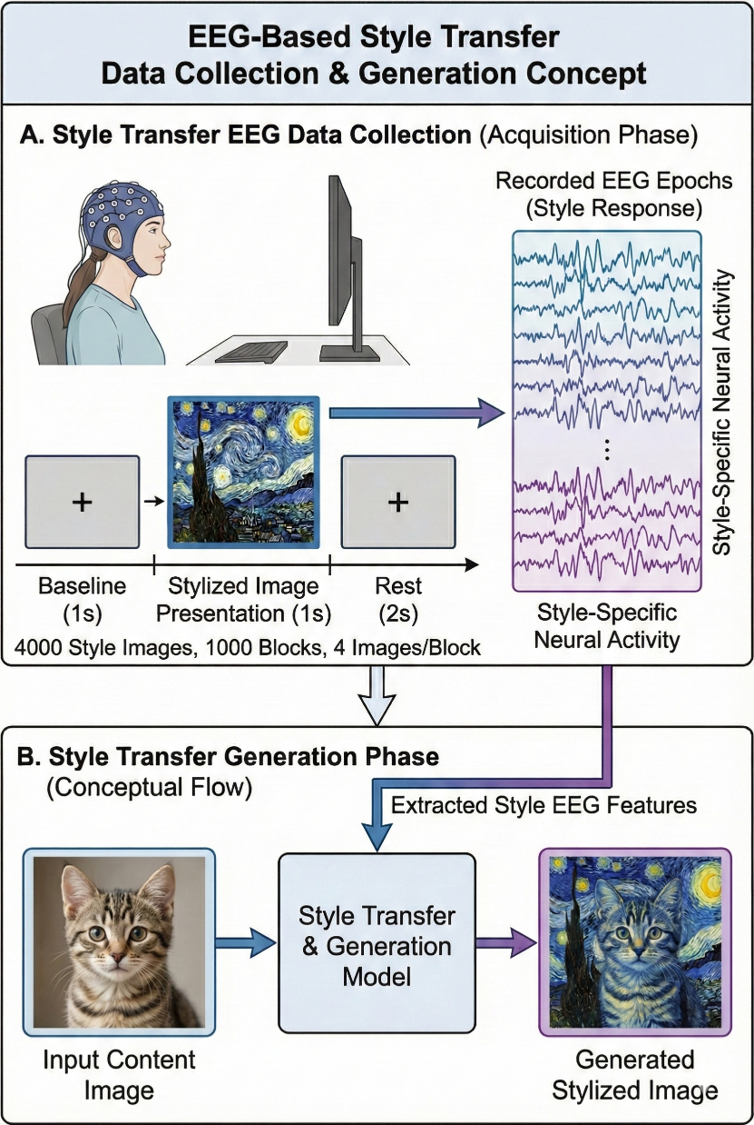}
  \caption{Experimental Environment for EEG-Style Dataset Acquisition.}
  \label{fig:exp}
\end{figure}

\textbf{Acquisition \& Synchronization.} The experimental setup for EEG data acquisition from participants is shown in Fig.\ref{fig:exp}. Scalp EEG was recorded using a standard international 10–20 electrode placement system, covering frontal, central, parietal, temporal, and occipital regions. Stimulus onset markers were sent from the stimulus presentation computer to the EEG acquisition system via a USB interface to provide precise digital triggers for each trial. In addition, a photodiode was attached to the display to measure the actual luminance change on the screen, enabling offline verification and correction of any timing discrepancies between the digital triggers and the true visual stimulus onset.

\textbf{Preprocessing \& Rationale.} 
The continuous EEG was band-pass filtered between 1 and 80 Hz. The high-pass cutoff at 1 Hz suppresses very slow drifts caused by electrode polarization, sweat, or skin potentials, while preserving low-frequency neural activity that is informative for cognitive state estimation. The low-pass cutoff at 80 Hz attenuates high-frequency muscle artifacts and hardware noise, yet retains the conventional EEG frequency bands: delta (1–4 Hz), theta (4–8 Hz), alpha (8–13 Hz), beta (13–30 Hz), and low gamma (30–80 Hz). Among these bands, theta and alpha oscillations are particularly sensitive to fluctuations in attention and mental workload, whereas beta and low-gamma activity are associated with task-related sensorimotor and higher-order visual processing, which are highly relevant for the EEG2IMAGE generation task.

\section{Additional Experiment Details}
\subsection{Definition of Style Descriptions}

\begin{figure}[!ht]
  \centering
  \includegraphics[width=\columnwidth]{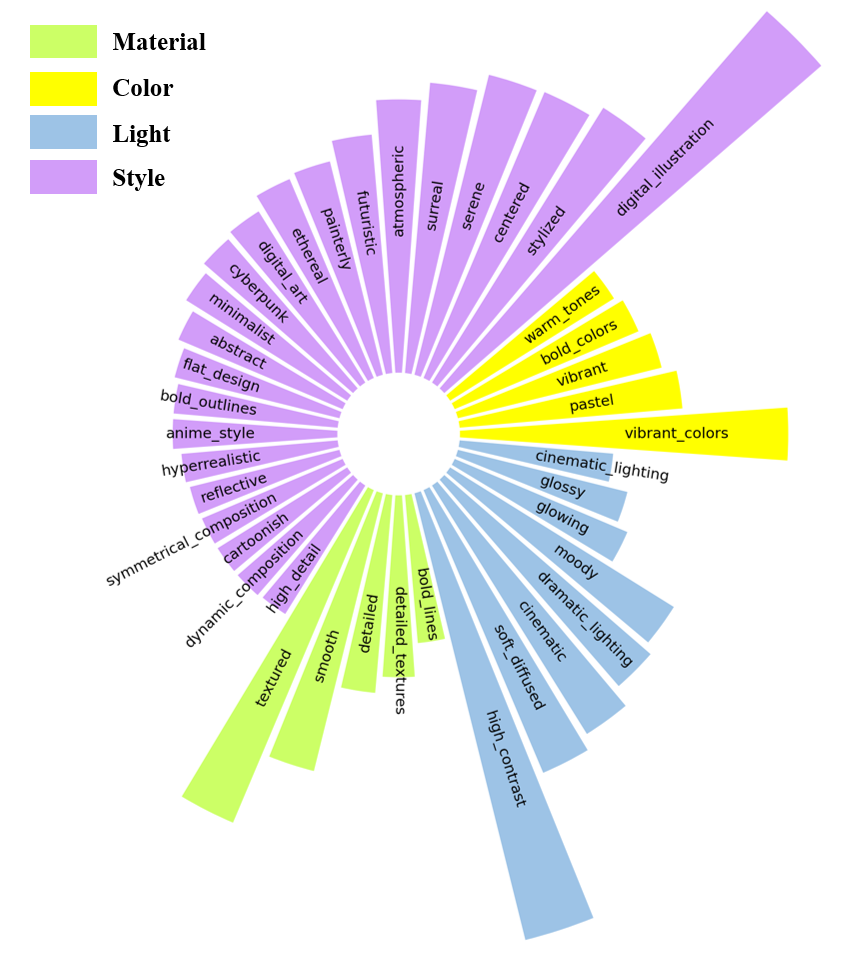}
  \caption{Top-40 Frequent Style Descriptors Generated by Qwen for Image Descriptions.}
  \label{fig:style-text}
\end{figure}

To investigate the discrepancy between EEG signals and textual descriptions, as well as their potential complementarity, we employ the Qwen3-VL-32B\cite{qwen3}, running on an NVIDIA A100 GPU, to summarize the visual style of each image in our dataset. Specifically, for every image the model is prompted to produce a short list of style-related keywords. We then compute the term-frequency statistics over all generated keywords and visualize the 40 most frequent terms, as illustrated in Fig. \ref{fig:style-text}

\begin{table*}[t]
\centering
\caption{Attention and EEG guided stylization scores of individual subject.}
\label{tab:subjects}
\begin{tabular}{cccccccccc}
\toprule
Subject & Gender & Age & \#Samples & Attention & CLIP-I$\uparrow$ & DINO$\uparrow$ & PSNR$\uparrow$ & SSIM$\uparrow$ & PDist $\downarrow$\\
\midrule
1  & Male & 25 & 4000 & 0.8095 & 0.7951 & 0.5278 & 9.99 & 0.2908 & 0.6643 \\
2  & Male & 21 & 4000 & 0.7142 & 0.7973 & 0.5253 & 9.93 & 0.2883 & 0.6677 \\
3  & Male & 22 & 4000 & 1.2475 & 0.7971 & 0.5331 & 9.94 & 0.2869 & 0.6647 \\
4  & Female & 25 & 4000 & 1.2009 & 0.7986 & 0.5280 & 10.02 & 0.2867 & 0.6660 \\
5  & Female & 25 & 4000 & 1.5208 & 0.7991 & 0.5284 & 9.95 & 0.2902 & 0.6673 \\

\bottomrule
\end{tabular}
\end{table*}

\subsection{Evaluation Metrics}
\label{sec:evaluation-metrics}
To comprehensively assess the quality of generated images, we employ a diverse suite of quantitative evaluation metrics covering pixel fidelity, perceptual similarity, semantic consistency, and distributional alignment. Pixel-level reconstruction quality is measured through L1 and L2 distances, as well as PSNR and SSIM, which capture low-level correspondence to ground-truth images. To evaluate perceptual similarity in deep feature space, we report LPIPS and PDist, both of which reflect human-aligned perceptual differences beyond raw pixel errors. Semantic consistency is assessed using CLIP-I, DINO, and CLIP-T scores, which measure alignment in pretrained vision–language or vision-only embedding spaces. Finally, we quantify realism and diversity of generated samples using Inception Score (IS), Fréchet Inception Distance (FID), and classification accuracy (ACC). Together, these metrics provide a comprehensive evaluation across multiple dimensions of image quality.

\textbf{L1 Distance.}
We report the mean absolute error (L1 distance) between generated and ground-truth images, averaged over all pixels. Lower L1 indicates better pixel-level reconstruction quality\cite{l1}.

\textbf{L2 Distance.}
We measure the mean squared error (L2 distance) between generated and ground-truth images. Lower L2 values correspond to closer alignment to the reference images\cite{l2}.

\textbf{CLIP-I Score.}
We compute CLIP-I as the average cosine similarity between CLIP image embeddings of generated and ground-truth images. Higher CLIP-I indicates better semantic consistency at the image level\cite{clip-i}.

\textbf{DINO Score.}
We use DINO features to compute the cosine similarity between generated and reference images. Higher DINO scores reflect better preservation of high-level semantic content\cite{dino}.

\textbf{CLIP-T Score.}
We evaluate text--image alignment by computing the cosine similarity between CLIP image embeddings of generated images and CLIP text embeddings of the corresponding prompts. Higher CLIP-T scores indicate better faithfulness to the textual description\cite{clip-t}.

\textbf{IS Score (Inception Score).}
We report the Inception Score computed with a pretrained Inception network. Higher IS indicates that generated images are both realistic and diverse\cite{is}.

\textbf{FID Score (Fr\'echet Inception Distance).}
We compute FID between Inception features of real and generated images. Lower FID means the generated image distribution is closer to the real data distribution\cite{fid}.

\textbf{ACC (Accuracy).}
We measure classification accuracy (ACC) by applying a fixed classifier to generated images and comparing predictions with ground-truth labels. Higher ACC indicates that the generated images preserve discriminative semantic information\cite{acc}.

\textbf{LPIPS.}
We use LPIPS to measure perceptual distance between generated and reference images in a deep feature space. Lower LPIPS values indicate higher perceptual similarity\cite{lpips}.

\textbf{PSNR (Peak Signal-to-Noise Ratio).}
PSNR is computed from the mean squared error between generated and ground-truth images. Higher PSNR corresponds to better reconstruction fidelity and less distortion\cite{PSNR}.

\textbf{SSIM (Structural Similarity Index).}
We compute SSIM between generated and reference images to assess structural similarity. Higher SSIM values indicate better preservation of image structure and perceived quality\cite{ssim}.

\textbf{PDist (Perceptual Distance).}
We report a perceptual distance (PDist) defined as the L2 distance between deep feature embeddings of generated and ground-truth images. Lower PDist indicates that generated images are closer to the references in feature space\cite{pdist}.

\section{Supplementary Case Studies}
\begin{figure*}[t]
  \centering
  % trim 的顺序是：left bottom right top；建议用 mm 为单位
  \includegraphics[width=\textwidth,trim=0mm 0mm 0mm 0mm,clip]{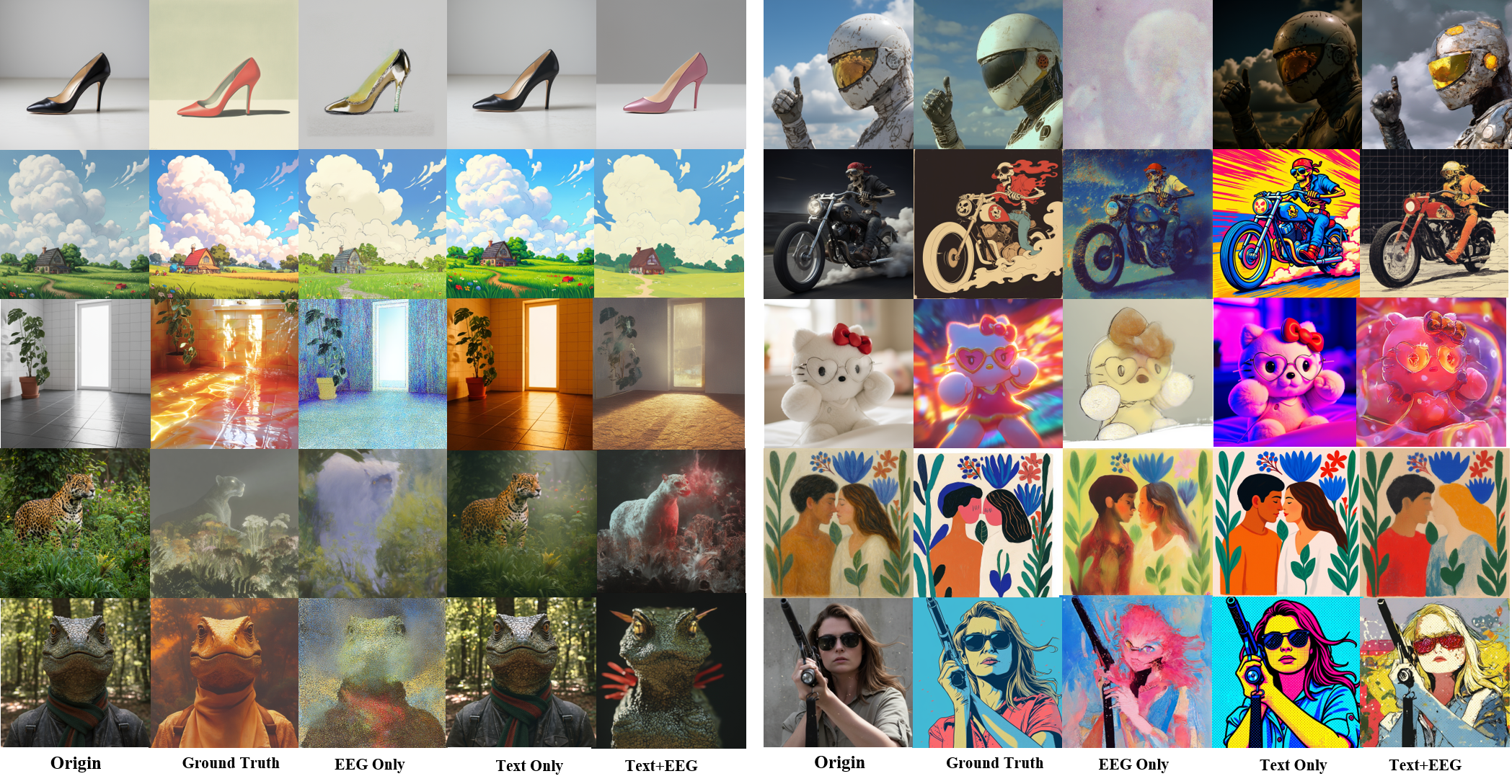}
  \caption{Qualitative results of Uni‑Neur2Img on the EEG‑Style dataset, comparing the original image, the ground‑truth target, and images generated under three input conditions: EEG Only, Text Only, and joint Text+EEG.}
  \label{fig:s2.2}
\end{figure*}

\begin{figure*}[t]
  \centering
  % trim 的顺序是：left bottom right top；建议用 mm 为单位
  \includegraphics[width=\textwidth,trim=0mm 0mm 0mm 0mm,clip]{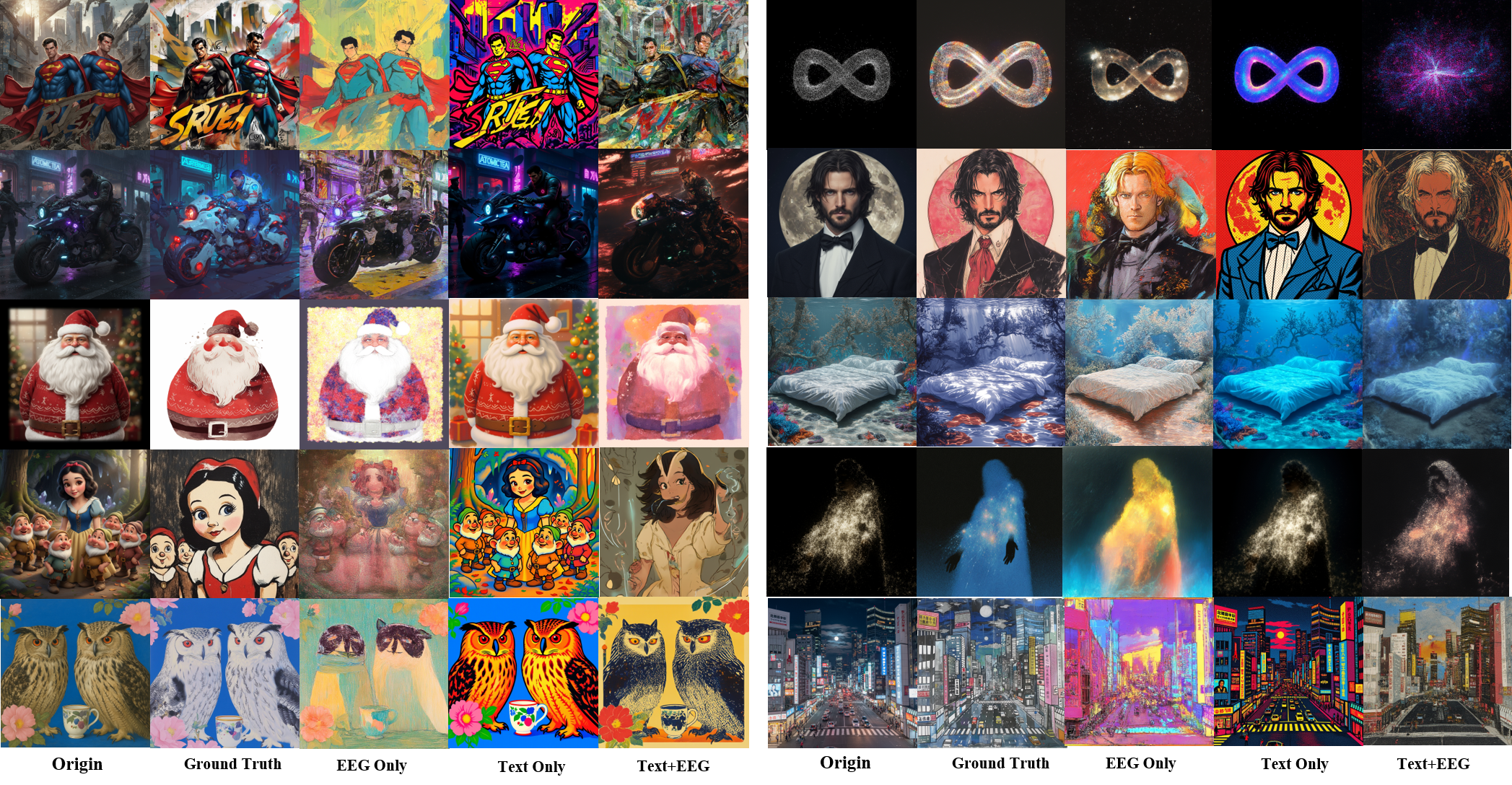}
  \caption{Qualitative results of Uni‑Neur2Img on the EEG‑Style dataset, comparing the original image, the ground‑truth target, and images generated under three input conditions: EEG Only, Text Only, and joint Text+EEG.}
  \label{fig:s2.3}
\end{figure*}

\textbf{Extended Qualitative Analysis on Style Transfer.}
Figures \ref{fig:s2.2} and \ref{fig:s2.3} provide a comprehensive visualization of Uni-Neur2Img's performance across a broad spectrum of artistic styles, ranging from photorealistic rendering and anime aesthetics to abstract pop art and complex mosaic patterns. These extended examples further illuminate the distinct roles of neural and textual modalities in the stylization process:

\begin{itemize}
    \item \textbf{Diversity of Style Representation:} As shown in Figure \ref{fig:s2.2}, our model successfully handles diverse material properties and lighting conditions. For instance, in the \textit{Hello Kitty} example (Row 3), the target style involves complex glowing lighting and semi-transparent materials. While the \textit{Text Only} baseline generates a generic plush toy, the \textit{EEG Only} result captures the correct pink neon atmosphere despite structural loss. The combined \textit{Text+EEG} output perfectly reconstructs the translucent, glowing aesthetic while maintaining the object's geometry. Similarly, for the landscape example (Row 2), the joint model captures the specific ``anime-style'' cloud fluffiness and saturation that the textual prompt alone fails to reproduce.

    \item \textbf{Handling High-Frequency Artistic Textures:} Figure \ref{fig:s2.3} demonstrates the model's capability in transferring intricate artistic textures. A prime example is the \textit{Snow White} case (Row 3), where the target style features a psychedelic, mosaic-like color distribution. The \textit{Text Only} variant produces a standard cartoon output, missing the unique artistic strokes. In contrast, the \textit{Text+EEG} result faithfully maps the complex color patches and high-frequency textures from the neural signals onto the semantic structure, creating a stylized image that is visually consistent with the ground truth.
    
    \item \textbf{Abstract and Stylized Composition:} In highly abstract scenarios, such as the \textit{Superman} pop-art example (Figure \ref{fig:s2.3}, Row 1) or the \textit{John Wick} illustrative portrait (Figure \ref{fig:s2.3}, Row 2), EEG signals prove critical in determining the background composition and color grading. The \textit{Text+EEG} model correctly renders the chaotic graffiti background for Superman and the specific red sun motif for John Wick, details that are inherently difficult to describe via text prompts alone.
\end{itemize}

These supplementary visuals strongly support our hypothesis: while text ensures semantic consistency, EEG signals act as a powerful descriptor for mid- and low-level stylistic attributes—including lighting, material texture, and artistic strokes—enabling high-fidelity style transfer that neither modality could achieve independently.

\section{Broader Societal Impacts}

This work explores image generation, editing, and stylization directly conditioned on non‑invasive neural signals such as EEG, using a unified diffusion‑transformer framework and a new EEG‑Style dataset that links multi‑subject EEG recordings with diverse visual styles.
 On the positive side, such techniques may enable more accessible creative tools for users with motor or language impairments, who may find it difficult to author detailed textual prompts but can still provide reliable neural responses to visual content. Neural‑driven interfaces could also support assistive communication, neuroadaptive user interfaces, and scientific studies of how the human brain encodes visual style, color, and structure. At the same time, the ability to decode aesthetic preferences or visually evoked responses from brain activity raises important concerns about privacy and potential misuse. In principle, similar methods could be repurposed for covert preference profiling, neuromarketing, or other forms of unwarranted monitoring of users’ mental states. Our current system is trained on small, controlled laboratory datasets with stylized images and healthy adult volunteers and is far from real‑world deployment,
 but we believe future work in this direction should incorporate explicit consent mechanisms, on‑device or privacy‑preserving training, and user control over data collection and retention. Dataset composition is another relevant factor: EEG‑Style primarily reflects young adults viewing Midjourney‑style imagery.
 which may bias downstream tools toward particular aesthetic conventions. Broader societal deployment should therefore be accompanied by more diverse data collection, transparency about capabilities and limitations, and robust safeguards against manipulative or non‑consensual use.

\section{Safety and Ethics}
All new EEG data in the EEG‑Style dataset were collected from healthy adult volunteers after full explanation of the procedure, written informed consent, financial compensation, and approval by the institutional ethics committee.
 The stimuli consist of stylized images sourced from an online generative gallery rather than photographs of the participants themselves, and the recorded signals are standard 64‑channel scalp EEG with additional ECG/EOG channels and basic preprocessing to remove noise.
 In addition to this new corpus, our experiments rely on existing public datasets (EEGCVPR40 and LoongX) that were collected under their respective ethical protocols.
 Across all experiments, we restrict ourselves to visually evoked activity and do not attempt to infer clinical status, personality traits, or other sensitive attributes from neural data. The learned models are used only for offline research on image reconstruction, editing, and stylization; no automated decisions about participants or users are made based on their brain activity. Nevertheless, EEG signals are inherently sensitive and may contain biometric or cognitive fingerprints. For any future release or deployment, we recommend treating neural recordings as protected personal data, applying strong de‑identification and secure storage, and accompanying the models and datasets with usage terms that explicitly prohibit surveillance, coercive monitoring, or other high‑risk applications. Finally, as with other generative image models, our framework could be misused to create deceptive or harmful content; responsible use should therefore pair such models with content‑filtering, provenance tracking, and clear communication about their synthetic nature.

\end{document}